\documentclass[twocolumn,10pt,cleanfoot]{asme2ej}

\usepackage[graphicx]{realboxes} 
\usepackage{amsfonts}
\usepackage{url}
\usepackage{enumitem}
\usepackage{float}
\usepackage{multirow}
\usepackage[table,xcdraw]{xcolor}
\usepackage{bm}
\usepackage{soul}
\usepackage{comment}
\usepackage{amsmath}

\title{Image2CADSeq: Computer-Aided Design Sequence and Knowledge Inference from Product Images}

\author{Xingang Li, Zhenghui Sha\thanks{Corresponding author.}
    \affiliation{
	Walker Department of Mechanical Engineering\\
	The University of Texas at Austin\\
	Austin, Texas 78712\\
    Email: zsha@austin.utexas.edu
    }	
}

\begin{document}

\maketitle    


\begin{abstract}
\textbf{Abstract:} Computer-aided design (CAD) tools empower designers to design and modify 3D models through a series of CAD operations, commonly referred to as a CAD sequence. In scenarios where digital CAD files are not accessible, reverse engineering (RE) has been used to reconstruct 3D CAD models. Recent advances have seen the rise of data-driven approaches for RE, with a primary focus on converting 3D data, such as point clouds, into 3D models in boundary representation (B-rep) format. However, obtaining 3D data poses significant challenges, and B-rep models do not reveal knowledge about the 3D modeling process of designs.
To this end, our research introduces a novel data-driven approach with an Image2CADSeq neural network model. This model aims to reverse engineer CAD models by processing images as input and generating CAD sequences. These sequences can then be translated into B-rep models using a solid modeling kernel. Unlike B-rep models, CAD sequences offer enhanced flexibility to modify individual steps of model creation, providing a deeper understanding of the construction process of CAD models.
To quantitatively and rigorously evaluate the predictive performance of the Image2CADSeq model, we have developed a multi-level evaluation framework for model assessment. The model was trained on a specially synthesized dataset, and various network architectures were explored to optimize the performance. The experimental and validation results show great potential for the model in generating CAD sequences from 2D image data.
\end{abstract}

\section{INTRODUCTION}
\label{INTRODUCTION}
Computer-aided design (CAD) systems can significantly reduce design time by avoiding the need for traditionally required labor-intensive manual drawings \cite{ROSATO2003Plastics}. Contemporary CAD systems such as Fusion 360, SOLIDWORKS, and OnShape enable designers to create and modify CAD models \footnote{CAD models are structured, parametric, or operation-based 2D or 3D design, allowing for a high level of control and flexibility in the design process. They are particularly suitable for industrial and engineering designs, where precision and the ability to easily modify designs are crucial. There are two main types of CAD models: 1) constructed solid geometry (CSG) and 2) parametric CAD models including CAD sequence data and boundary representation (B-rep) models. In contrast, discrete 3D representations, such as meshes and point clouds, are more static and less flexible in terms of parametric editing and design exploration \cite{wu2021deepcad, para2021sketchgen}.} through a sequence of CAD operations.
However, in certain scenarios, the CAD model of a product may not be readily available due to various factors, including outdated documentation, lack of digital records, and commercial reasons. Reverse engineering (RE) is employed to overcome these obstacles, utilizing measurement and analysis tools to reconstruct CAD models \cite{varady1997reverse,buonamici2018reverse}. 

Integrating RE with CAD systems can not only allow designers to leverage the advantages of existing products while incorporating their own innovative ideas and improvements but can also be used for design knowledge restoration and management. 
However, the traditional RE process faces two major limitations. First, it focuses on reconstructing 3D models rather than CAD sequences. Compared to 3D models, a CAD sequence provides access to the historical construction process and associated design knowledge and it facilitates geometry modification using parametric modeling. Second, the process has been performed primarily manually, making it labor-intensive and time-consuming. Recently, researchers have explored data-driven methods, such as converting 3D point clouds \cite{uy2022point2cyl, ren2022extrudenet} or voxels \cite{lambourne2022reconstructing, li2023secad} into CAD models. Nevertheless, these 3D input data are often challenging to acquire due to inaccessibility and unavailability. 3D scanning could be a solution, yet quality is often unsatisfactory and cost is an unavoidable factor to consider when acquiring specialized equipment and expertise.

Compared to point clouds or voxels, images are easier to acquire given the popularity of mobile devices. 
Thus, our question arises: How can we reverse engineer CAD sequences directly from 2D images in supporting designers to interpret and edit CAD models during the design and modeling process? After a thorough literature review, we realize that there is a scarcity of research exploring answers to this question. Therefore, our objective is to develop a data-driven approach that can generate a sequence of CAD operations based on a single image (referred to as ``Image2CADSeq" hereafter for brevity). 

The contributions of the proposed approach are summarized as follows.
\begin{enumerate}
    \item [a).] To the best of our knowledge, this study is the first attempt to predict a sequence of CAD operations given a single image input (i.e., single-view image to CAD sequence prediction). We developed a target-embedding variational autoencoder (TEVAE) architecture \cite{li2022predictive} to solve this problem. 
    The proposed approach has the potential to streamline the CAD design process, reducing the time and effort required to create 3D models from 2D sketches or photographs.
    \item [b).] We created a novel data synthesis pipeline based on the design grammars defined in the domain-specific language (DSL),  Fusion 360 Gallery. The pipeline can generate synthetic data that resemble real-world images and CAD models. It can also be used as a data augmentation method to improve the quality and quantity of existing training data sets, making the data more diverse and robust for training image2CADSeq models. 
    \item [c).] We developed a multi-level evaluation framework to assess the Image2CADSeq performance, encompassing three components: CAD sequences, 3D models, and the corresponding images. Specifically, the evaluation of the CAD sequence is performed at multiple levels and hierarchies (see Section \ref{sec:metrics} for details) to quantitatively assess the predictive performance of the proposed model architectures.  
\end{enumerate}

We anticipate that the proposed approach has the potential to revolutionize existing CAD systems by making the CAD model reconstruction process more accessible. This would enable both experienced and novice designers to actively contribute to the design, promoting design collaboration and design education. Moreover, it has the potential to provide a unique pathway to involve end users in the design process, promoting design democratization.

The remainder of this paper is organized as follows. In Section \ref{literature review} and \ref{tech_background}, we provide an overview of the background related to data-driven 2D-to-3D generation and technical background about the applied techniques. Section \ref{sec:methodology} outlines the methodology in the development of our Image2CADSeq model. Subsequently, Sections \ref{IMPLEMENTATION DETAILS AND RESULTS} and \ref{sec:discussion} present and analyze the experimental results, summarizing the primary findings and acknowledging limitations. Conclusions and closing remarks are presented in Section \ref{sec:CONCLUSION}, where we present key insights and suggest potential directions for future research.

\section{LITERATURE REVIEW}
\label{literature review}
In this section, we first provide a background of deep learning techniques applied to 2D-to-3D generation using discrete 3D representations, such as voxels and meshes, as well as the generation of CAD models using constructive solid geometry (CSG) and 2D sketches. Following this introduction, we then present a review of deep learning methods specifically tailored for 3D parametric CAD models, which are most relevant to our work. 

\subsection{RESEARCH ON 2D-TO-3D GENERATION}
Our research is related to the domain of 2D-to-3D generation, led by advances in computer graphics and computer vision. For an in-depth understanding of the current developments and challenges in this area, we direct readers to a comprehensive review \cite{shi2022deep}. The use of discrete 3D representations, such as voxels, point clouds, and meshes, has been prevalent. Despite their widespread use, these representations often focused on generating visually appealing objects without necessarily considering their engineering aspects, such as dimensions, engineering performance, and compatibility with engineering software \cite{li2023deep}. This mismatch hinders the seamless integration with downstream applications, such as editing and engineering analysis of synthesized 3D shapes, underlining the necessity for adopting CAD-specific data formats in our research. 

Constructive solid geometry (CSG) is one of the fundamental methods for creating CAD models. It applies Boolean operations (e.g., union, intersection, and subtraction) to basic geometric shapes (primitives), such as cuboids, spheres, and cylinders. CSG is known for its lightweight structure, which allows for easy modifications by altering the parameters of these primitives and their spatial transformations. There have been various studies on the deep learning methods of CAD model generation using CSG \cite{ren2021CSG, sharma2017CSGNet, sharma2019Neural, kania2020ucsg}. Despite its merits, CSG lacks the versatility used in contemporary CAD tools that utilize parametric modeling.

Parametric CAD models start as 2D sketches comprising geometric primitives (e.g., line segments and arcs) with explicit constraints, such as coincidence and perpendicularity, establishing the foundation for 3D construction operations (e.g., extrusion and revolution).
The relevant deep learning methods of parametric CAD models include 2D engineering sketch and 3D model generation and reconstruction. 
There has been a series of studies recently \cite{willis2021engineering, seff2021Vitruvion, ganin2021computer, para2021sketchgen, yang2022Discovering} dedicated to the generation of CAD sketches through the application of deep learning approaches. The emphasis in these works is on generating 2D layouts rather than dealing with the generation of 3D components. We will be focused on introducing deep learning methods for 3D CAD model generation and reconstruction since they are more relevant to our work. 

\subsection{DEEP LEARNING OF PARAMETRIC 3D CAD MODELS}

Boundary representation (B-rep) format is the standard format for representing 3D shapes in CAD, which defines objects based on their boundary surfaces, edges, and vertices connected through specific topology. Numerous learning-based approaches have emerged for the generation of parametric curves \cite{wang2020pie} and surfaces \cite{sharma2020parsenet}. In addition to curve or surface generation, Smirnov et al. \cite{smirnov2020learning} introduced a generative model for creating topology that combines parametric curves and surfaces to create solid models, depending on predefined topological templates. In addition, various methods have been proposed to enable the direct generation of B-rep models with arbitrary topology \cite{guo2022complexgen, jayaraman2022solidgen, wang2022neural}. Different from these works, our focus lies in generating CAD sequences that can be translated into B-rep models using a solid modeling kernel, such as Fusion 360 CAD software. 

Significant progress has been made in the generation of CAD sequences for the reconstruction of 3D models particularly through \textit{Sketch-and-Extrude} modeling operations. Recently, there have been methods \cite{wu2021deepcad, xu2022skexgen} for generative models specifically designed for the unconditional generation of CAD sequences. These models aim to autonomously create CAD sequences without relying on specific conditions or inputs. Specifically, Wu et al. \cite{wu2021deepcad} presented the first generative model, DeepCAD, that learns from sequences of CAD modeling operations to produce editable CAD designs. By drawing an analogy between CAD operations and natural language, the authors propose to utilize a transformer \cite{vaswani2017attention} architecture aiming to leverage the capabilities of transformer models in understanding and generating sequences, adapting them to the context of CAD design operations.

Generative models indeed serve as valuable tools for randomly generating a multitude of designs, offering inspiration and exploration of diverse possibilities. However, these models lack the capability to directly incorporate designers' intent into the generation process. Consequently, the designs generated can deviate from the expectations or specific requirements of the designers. This discrepancy highlights the need for mechanisms that allow designers to guide or influence the output, ensuring that the generated designs align more closely with their intent and preferences.
To that end, several methods have been introduced to allow the CAD sequence generation given the target of B-rep models \cite{willis2021fusion, xu2021inferring}, voxels \cite{lambourne2022reconstructing, li2023secad}, point clouds \cite{uy2022point2cyl, ren2022extrudenet}, and sketches \cite{li2020sketch2cad, li2022free2cad}. 
Particularly, Fusion 360 Gym \cite{willis2021fusion} was developed to reconstruct a CAD model given a B-Rep model, utilizing a face-extrusion technique that relies on existing planar faces within the B-Rep model. However, despite the potential for CAD sequence generation, the face-extrusion method differs significantly from the more natural sketch-extrusion method commonly used by human designers. Moreover, this technique is ineffective when confronted with a lack of available planar or profile data in the input data, such as images.

Our work aims to fill a research gap in the existing literature by focusing on the task of generating CAD sequences from images. In particular, we expand upon the transformer-based autoencoder initially introduced in DeepCAD \cite{wu2021deepcad} and convert it into a TEVAE architecture developed in our previous work \cite{li2022predictive}. In the case study, we apply the domain-specific language of Fusion 360 Gym \cite{willis2021fusion} that demonstrates our approach to predicting CAD sequences that involve \textit{Sketch-and-Extrude} operations. 

\section{TECHNICAL BACKGROUND}
\label{tech_background}
An autoencoder (AE) is a type of neural network that aims to learn a compressed representation of input data \cite{hinton2006reducing}. It consists of two main parts: an encoder and a decoder. The encoder compresses the input into a lower-dimensional latent space, while the decoder reconstructs the input data from this compressed representation. The goal is to minimize the difference between the original input and its reconstruction, leading to efficient data encoding. 
Variational Autoencoders (VAEs) \cite{kingma2014autoencoding} extend traditional AEs by introducing a probabilistic way to the encoding process. This probabilistic approach allows VAEs not only to reconstruct input data but also to generate new data that is similar to the input. 
Autoencoders have been applied to uncover the useful underlying structures of data which is typically known as representation learning \cite{bengio2013representation}. 

Representation learning is a set of techniques in machine learning that automatically discovers the representations with reduced dimensionality from raw data for downstream tasks, such as regression or classification \cite{khastavaneh2019representation}. While VAEs are particularly known as generative models for their effectiveness in generating complex data, they offer a more robust, regularized, and probabilistic approach to representation learning compared to traditional AEs \cite{li2022predictive, gomari2022variational}. 

Although most research has focused on employing AEs and VAEs in unsupervised or semi-supervised scenarios, it is worth noting that autoencoders also demonstrate utility in supervised contexts \cite{jarrett2020target}. Specifically, incorporating an auxiliary feature-reconstruction task proves beneficial in enhancing supervised classification problems \cite{le2018supervised}, which are known as feature-embedding autoencoders (FEAs). Furthermore, Jarrett and Schaar \cite{jarrett2020target} propose target-embedding autoencoders (TEAs) and demonstrate their effectiveness theoretically and empirically. As implied by their names, TEAs focus on encoding the target's information into a latent space, while FEAs encode the feature's information. 

TEAs have been applied to various problems. Girdhar et al. \cite{girdhar2016learning} introduce a TL-embedding network, comprising a T-network (an autoencoder network for the horizontal bar and an encoder for the vertical bar) during training. Upon training completion of the T-network, it facilitates the derivation of the L-network, enabling the prediction of 3D voxel shapes from input images. A similar network architecture has also been applied for semantic image segmentation tasks \cite{mostajabi2018regularizing, dalca2018anatomical}. Drawing inspiration from these preceding studies, Li et al. \cite{li2022predictive} propose to use a VAE to replace the AE and form a target-embedding variational autoencoder (TEVAE) architecture, demonstrating its effectiveness in predictive and generative tasks for car and mug design examples. 

In this study, we constructed the Image2CADSeq neural network model by comparing both TEA and TEVAE architectures. Our objective is to assess their effectiveness in the prediction task of images to CAD sequences. The results revealed a significant superiority of the TEVAE architecture over the TEA architecture in terms of prediction performance as detailed in Section \ref{subsec:results}.

\begin{figure}
    \centering
    \includegraphics[width=\linewidth]{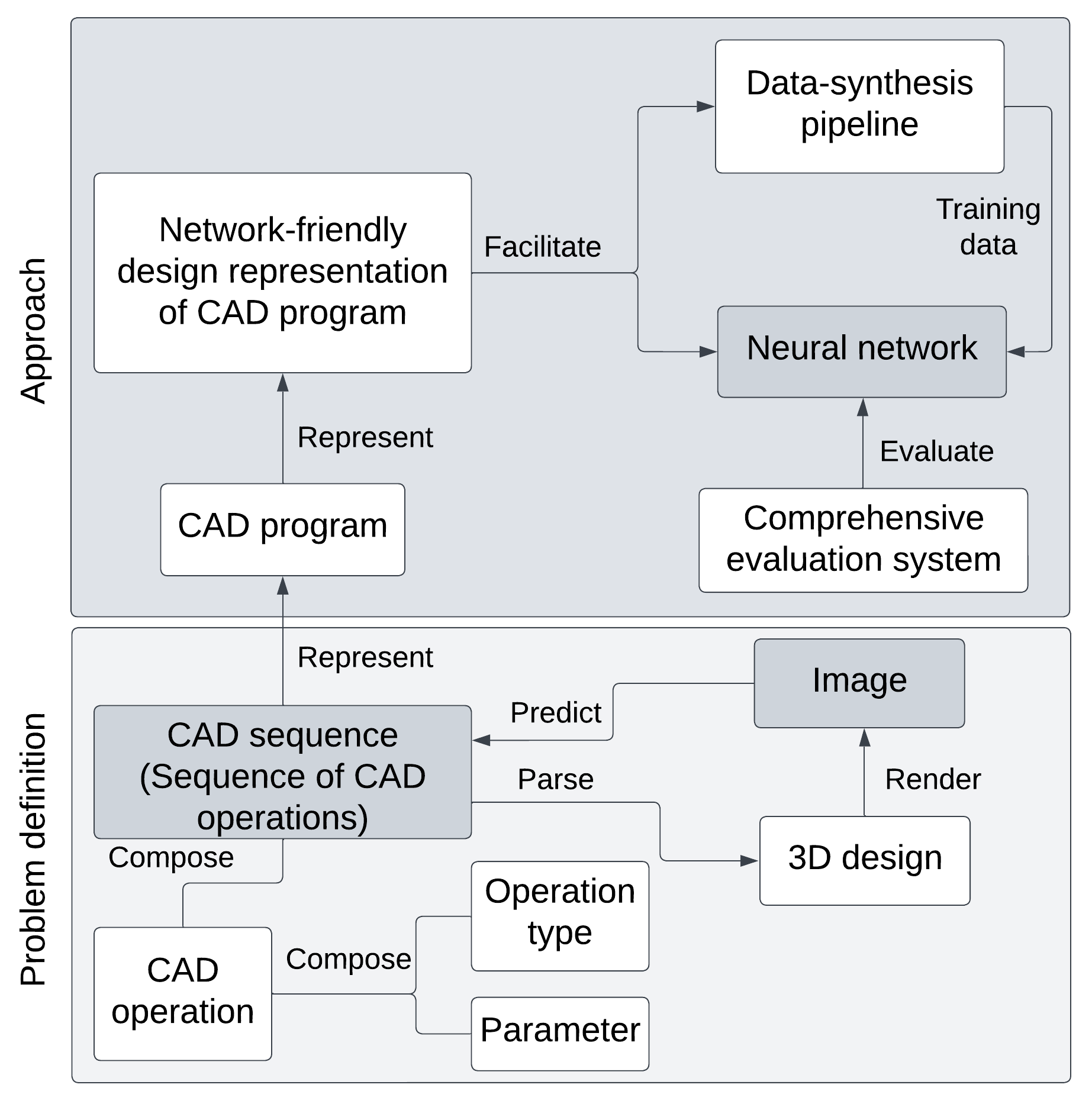}
    \caption{Approach overview. }
    \label{fig:approach_overview}
\end{figure}

\section{METHODOLOGY}
\label{sec:methodology}
The flowchart depicted in Figure \ref{fig:approach_overview} illustrates the proposed systematic approach to predicting CAD sequences given images, a process we refer to as Image2CADSeq. To effectively tackle this challenge, we initiate with a clear problem definition that divides the task into manageable components. Our objective is to harness the power of deep learning to predict a CAD sequence—a series of CAD operations characterized by specific operation types and their corresponding parameters—from an image. The image could be a rendering from a CAD model or a real-world photograph of a 3D object. 

Due to the intricate nature of CAD sequences, we employ a CAD program as a representational tool for CAD sequences. A CAD program enables designers to script their designs programmatically in a specialized scripting environment, such as the Fusion 360 API, FreeCAD API, or CADQuery. CAD programs can convert CAD sequences described in text into script language that is interpretable and executable by computers. 
To overcome the inherent lack of structured format in the CAD sequence data, the CAD program is then streamlined into a vectorized representation conducive to neural network processing. This representation can facilitate not only the development of our neural network's architecture but also the creation of a data-synthesis pipeline tasked with generating the training data for the neural network. In addition, given the complexity of the Image2CADSeq task, we develop a comprehensive evaluation system that rigorously assesses our neural network models' performance, thereby ensuring the reliability and accuracy of our approach. This holistic evaluation is crucial in refining the Image2CADSeq model and guiding its evolution to meet the demanding standards of CAD sequence prediction.

In this study, we employ a particular domain-specific language (DSL), namely Fusion 360 Gallery (abbreviated as Gallery for conciseness) \cite{willis2021fusion} for the CAD program, as a solid case to demonstrate our approach. 
Therefore, before the presentation of how we devise the vectorized design representation for the CAD sequence data, we provide an introduction to the Gallery DSL below.

\begin{table}
  \caption{Fusion 360 Gallery Domain Specific Language}
  \label{tab:gallery_dsl}
  \centering
  \includegraphics[width=\linewidth]{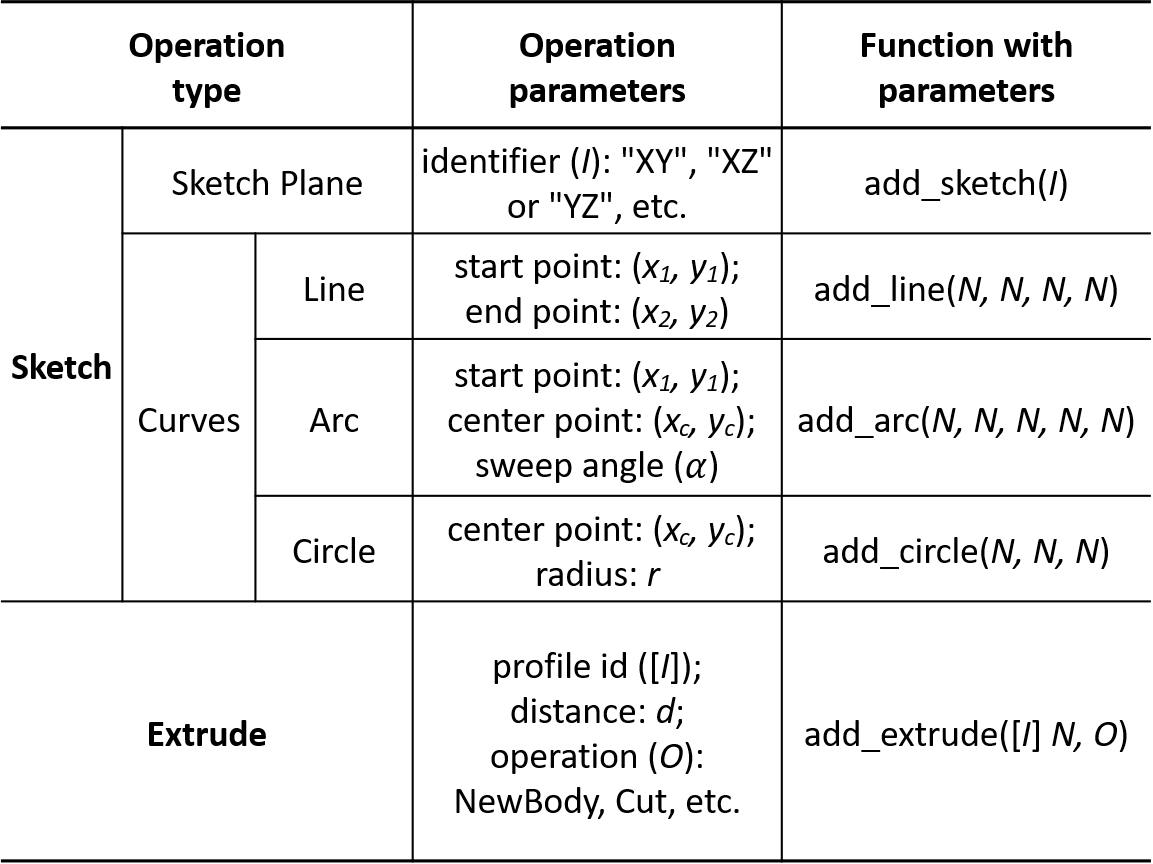}
\end{table}

\begin{table*}
  \caption{Comparison of the CAD programs for creating a cylinder with a base circle radius of 5 and a height of 10 using Fusion 360 Python API, Fusion 360 Gallery DSL, and the Simplified Gallery DSL. Note: Bullet numbers are used to indicate the effort that a designer would typically make to create a design.}
  \label{tab:comparison_DSLs}
  \centering
  \includegraphics[width=\textwidth]{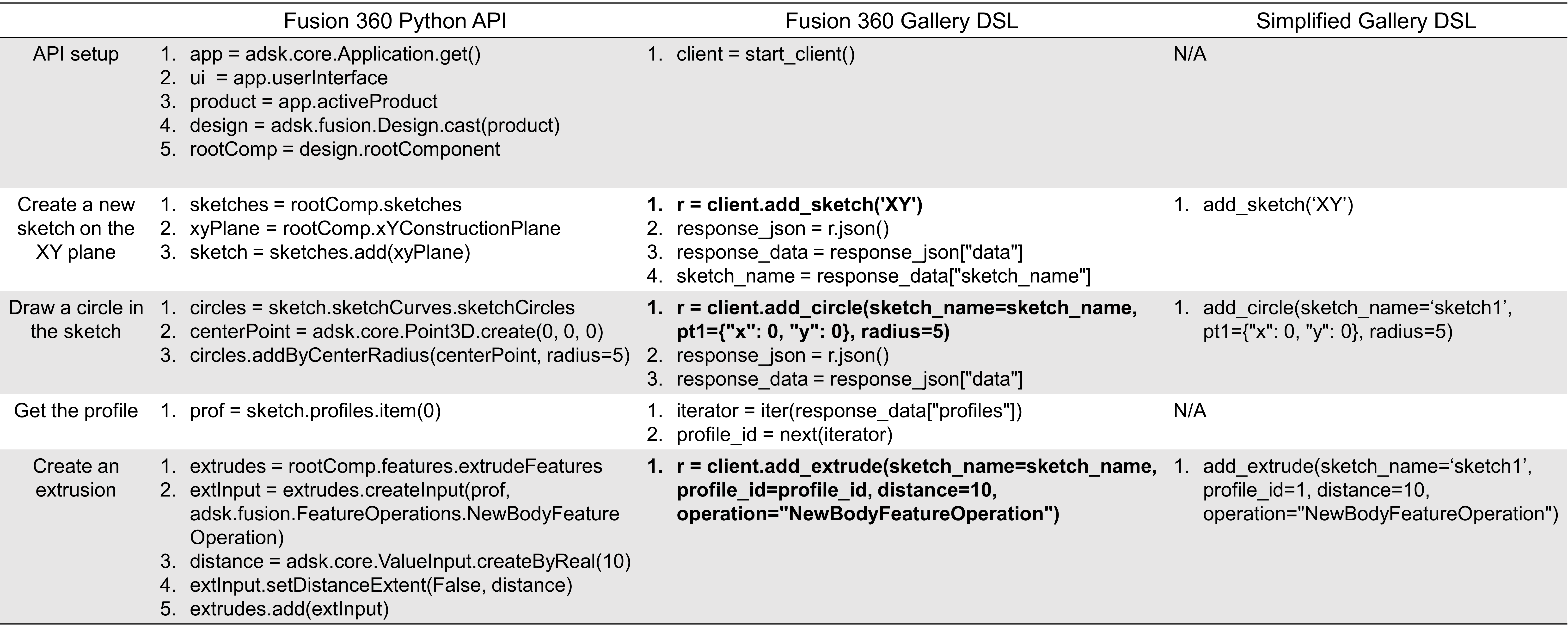}
\end{table*}

\subsection{FUSION 360 GALLERY DOMAIN SPECIFIC LANGUAGE}
\label{sec: gallery and gym}

Table \ref{tab:gallery_dsl} presents a summary of the core elements (i.e., CAD-related elements) in Gallery DSL, which enables the representation of a 2D/3D design as a CAD program, and Python is used to implement the CAD operations \cite{willis2021fusion}. 
Gallery DSL now supports two major types of CAD operations: \textit{Sketch} and \textit{Extrude}. 
Each CAD operation is decomposed into two fundamental components: the operation type and its corresponding parameters. These elements are analogously mirrored in the Gallery DSL as function names and their associated parameters.

A \textit{Sketch} operation includes the definition of a \textit{Sketch Plane} and \textit{Curves} on it. A \textit{Sketch Plane} can be created by the $add\_sketch(I)$ function, where $I$ is a plane identifier that can be specified from the three canonical planes "XY", "XZ", or "YZ" or other planar faces (e.g., the side face of a cube) present in the current geometry. The \textit{Sketch Plane} can then be used as the reference coordinate system in 2D for specifying the coordinates. 
A sequence of \textit{Curves}, including \textit{Line}, \textit{Arc}, or \textit{Circle}, can be drawn using $add\_line(N, N, N, N)$, $add\_arc(N, N, N, N, N)$, and $add\_circle(N, N, N)$, respectively, Where $N$ is a real number representing the required parameters for a particular operation. As two numbers are needed to define one point, \textit{Line} uses four numbers for start and endpoints; \textit{Arc} needs five numbers: start point, center point, and sweep angle; and \textit{Circle} is specified with three numbers: two for position and one for radius. Executing a \textit{Sketch} can result in enclosed regions, termed profiles in CAD language. An \textit{Extrude} operation can extrude a profile from 2D into 3D by using $add\_extrude(I, N, O)$, where $I$ is an identifier for the profile, and $N$ is a signed number defining the depth of extruding along the normal direction of the profile. The Boolean operation $(O)$ specifies the behavior of the extruded 3D volume, e.g., add to or subtract from other 3D bodies.

While Gallery DSL currently does not support certain objects, such as spheres and springs, it still covers a vast range of them by using expressive \textit{Sketch} and \textit{Extrude} operations with Boolean capability \cite{willis2021fusion}. Therefore, it is a good starting point for supporting learning-based methods \cite{willis2021fusion, wu2021deepcad}. In the future, the other CAD operations, such as \textit{Revolve}, \textit{Sweep}, and \textit{Fillet}, can be added to the Gallery DSL to expand its design grammar for a full-fledged CAD tool. In this study, we leverage the current status of the Gallery DSL by only considering the \textit{Sketch} and \textit{Extrude} operations.

Gallery DSL acts as a simplified interface to the more complex Fusion 360 Python API. In essence, it democratizes access to sophisticated CAD design through a more intuitive Python-based interface, effectively bridging the gap between complex CAD operations and the user’s ability to execute them efficiently. For instance, as demonstrated in Table \ref{tab:comparison_DSLs}, when creating a cylinder with a base circle radius of 5 and a height of 10, the CAD program using Gallery DSL requires only about two-thirds of the efforts needed with the Fusion 360 Python API and does not require sophisticated definitions of various variables. This makes the Gallery DSL code easier to operate, and more user-friendly and accessible. However, Gallery DSL still requires a substantial amount of coding efforts in non-CAD-related elements, beyond the core functions as listed in Table \ref{tab:gallery_dsl}. 

To further improve its readability and accessibility, we simplify the Gallery DSL by isolating its key CAD-related functions referred to as Simplified-Gallery DSL or Sim-Gallery DSL for briefness, as shown in Table 2. We create the Sim-Gallery DSL following the concept of parametric modeling. In parametric modeling, a design is a sequence of operations that progressively modify the current geometry of an object. This process can be well represented in the Sim-Gallery DSL by a series of pure CAD operation functions.
This simplification reduces the process to just three steps for creating a cylinder: \(add\_sketch(\cdot)\), \(add\_circle(\cdot)\), and \(add\_extrude(\cdot)\). Additionally, we have developed a parsing method in Python to convert this simplified version back to the standard Gallery DSL. This ensures compatibility with Fusion 360 CAD software, facilitating seamless integration and execution of the CAD programs written in the Sim-Gallery DSL. It can also facilitate the design of the vectorized representation for the CAD sequence data as introduced in Section \ref{sec: vector rep}.

\begin{table}
  \caption{Variables $t, I, x, y, \alpha, r, [I], d, O, s$ for the vectorized design representation of Gallery DSL}
  \label{tab:vec_rep}
  \centering
  \includegraphics[width=\linewidth]{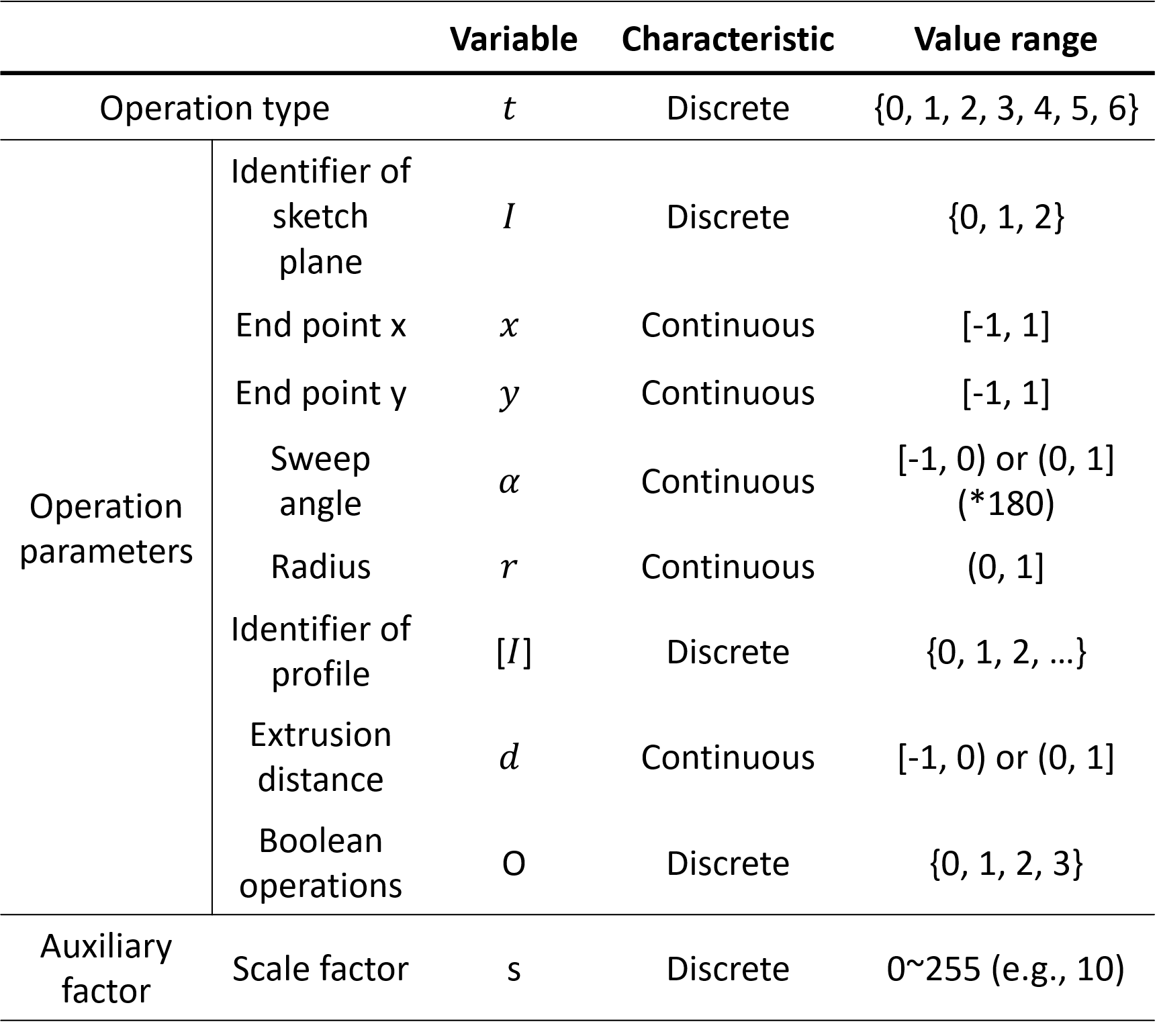}
\end{table}

\subsection{DESIGN REPRESENTATION OF CAD PROGRAMS}
\label{sec: vector rep}
A standardized design representation is essential for neural networks to effectively interpret CAD programs. Thus, it becomes crucial to devise an efficient method to represent each CAD operation and the entire CAD program. There are three major challenges:

\begin{enumerate}
    \item Diversity in CAD operations: Different CAD programs comprise varying numbers of operations.
    \item Variability in parameters: Different CAD operations involve different numbers of parameters.
    \item Type of parameters: Parameters can be either continuous or discrete values.
\end{enumerate}
 
To tackle these challenges, we propose to use a design representation with a unified data structure. 
We identified 10 variables (\(t, I, x, y, \alpha, r, [I], d, O, s\)) from the Sim-Gallery DSL, detailed in Table \ref{tab:vec_rep}. In what follows, we elaborate on our approach to handling these variables.

\begin{enumerate}
    \item[(1)] $t\in\{0, 1, 2, 3, 4, 5, 6\}$ represents the operation types with $0-4$ representing $add\_sketch$, $add\_line$, $add\_arc$, $add\_circle$, and $add\_extrude$. The values $5$ and $6$ are used to represent the start $(SOP)$ and the end $(EOP)$ of a CAD program, which are not typical CAD operations but are included for the learning process to indicate a complete CAD program as required by a transformer model \cite{vaswani2017attention, carlier2020deepsvg, wu2021deepcad}.  
    \item[(2)] $I\in\{0, 1, 2\}$ indicates the \textit{Sketch Plane} using one of the canonical planes: "XY", "XZ", or "YZ". 
    \item[(3,4)] $x$ and $y$ are the coordinates of the endpoint for \textit{Line} and \textit{Arc}, while they represent the center point when the operation type is \textit{Circle}. We excluded the start point required by \textit{Line} and \textit{Arc} from the design representation by obtaining it from the precedent curve to make sure all curves are connected one after another, making the vectorized representation more compact. There are two extra considerations for this setting: (i) If one curve has no precedent, we default its start point to the origin $(0, 0)$ when parsing the design representation. (ii) For \textit{Arc} that requires a center point instead of an endpoint, we calculate the coordinates for the center point based on its start point, endpoint, and sweep angle. 
    \item[(5)] $\alpha$ represents the sweep angle of an \textit{Arc}. 
    \item[(6)] $r$ is the radius of a \textit{Circle}. 
    \item[(7)] $[I]$ represents the profile index in the \textit{Sketch}. 
    \item[(8)] $d$ represents the signed distance of the depth for \textit{Extrude}. 
    \item[(9)] $O\in\{0, 1, 2, 3\}$ is used to indicate the Boolean operations: join, cut, intersect, or add, respectively.
    \item[(10)] $s$ is an auxiliary factor that can be used to scale a CAD model.
\end{enumerate}

\begin{figure*}
    \centering
    \includegraphics[width=0.8\linewidth]{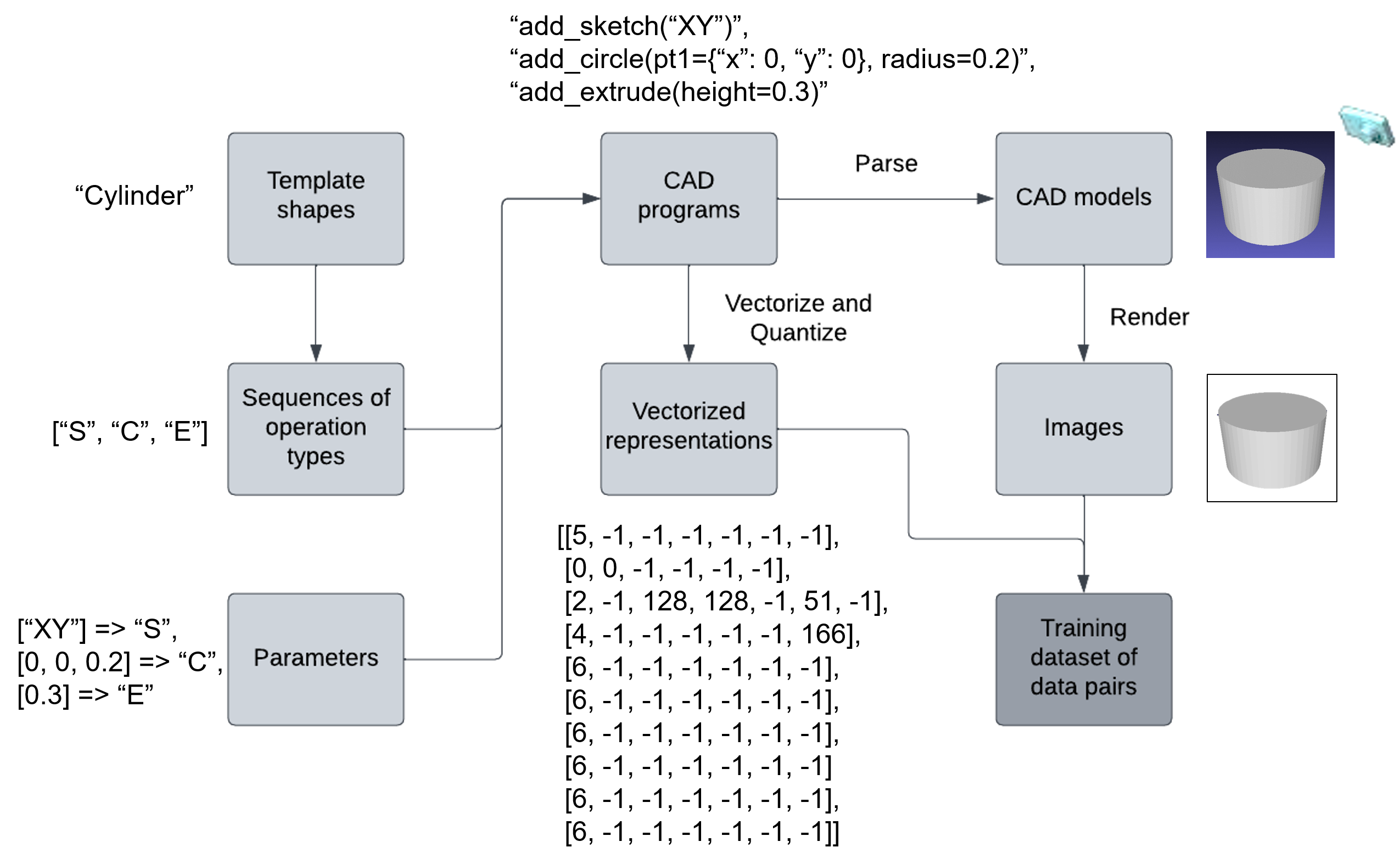}
    \caption{Synthesis pipeline for the training dataset of data pairs of image and vectorized CAD sequence, exemplified using a cylinder model}
    \label{fig:data_synthesis_diagram}
\end{figure*}

In addition, to standardize the treatment of both continuous and discrete parameters, inspired by \cite{carlier2020deepsvg, wu2021deepcad}, we discretize continuous parameters through quantization. This involves: (a) Confining continuous values to a subset of \([-1, 1]\) (e.g., \((0, 1]\) for radius and \([-1, 1]\) for endpoint x and y; (b) Dividing each range into 256 equal segments, enabling representation as 8-bit integers (i.e., \(0-255\)); (c) For the sweep angle (\(\alpha\)), we multiply it by \(180\) during interpretation; (d) Handling scale factor (\(s\)): Although the scale factor can be a non-negative continuous value, we limit it to 256 levels for consistency with other continuous values' quantization. Consequently, the 10 variables can encode both the operation type and its associated parameters. From the 10 variables, a fixed-dimensional vector can be formalized as a unified design representation for each CAD operation, and the unused parameters will be filled with values of $-1$. 
See Figure \ref{fig:data_synthesis_diagram} for an example. 

The subsequent consideration involves standardizing CAD programs of varying sizes (the number of CAD operations involved). For example, besides the start and end marks, $SOP$ and $EOP$, a cylinder can be created in three operations as shown in Table \ref{tab:comparison_DSLs}, while a triangular prism in five operations (\(add\_sketch(\cdot)\), \(add\_line(\cdot)\times 3\), and \(add\_extrude(\cdot)\)). To achieve a consistent data structure across all CAD programs, we introduce a treatment, called \textit{maximum program length}. Then, CAD programs shorter than this maximum length are extended by appending end marks (\textit{EOP}) until they reach the predetermined length. 

In this study, we use 7 variables to construct a 7-dimensional vector \([t, I, x, y, \alpha, r, d]\) for each CAD operation (i.e., one step/line in a CAD sequence). 
Additionally, we assign default values to the other three variables \([I], O\), and \(s\), setting them as \(0\), \(3\), and \(10\) correspondingly. Different variables can be selected which will influence the complexity of the data structure and thus the complexity of designs. In addition, the maximum length for CAD programs is set to 10. As a result, the design representation of a CAD program will be a matrix, namely, the feature matrix.
Mathematically, the feature matrix denoted as \(P\), is expressed as $P = \begin{bmatrix} \mathbf{o^1,} & \mathbf{o^2,} & \ldots & \mathbf{, o^{N_c}} \end{bmatrix}^T \in \mathbb{R}^{10\times7}$, 
where \(\mathbf{o^i}\in \mathbb{R}^7\) is a CAD operation vector, and \(N_c=10\) is the sequence length of the CAD program. Refer to Figure \ref{fig:data_synthesis_diagram} for an example of how a cylinder is converted to a feature matrix, including the quantization of its parameters as explained earlier. 

\begin{figure}
    \centering
    \includegraphics[width=0.9\linewidth]{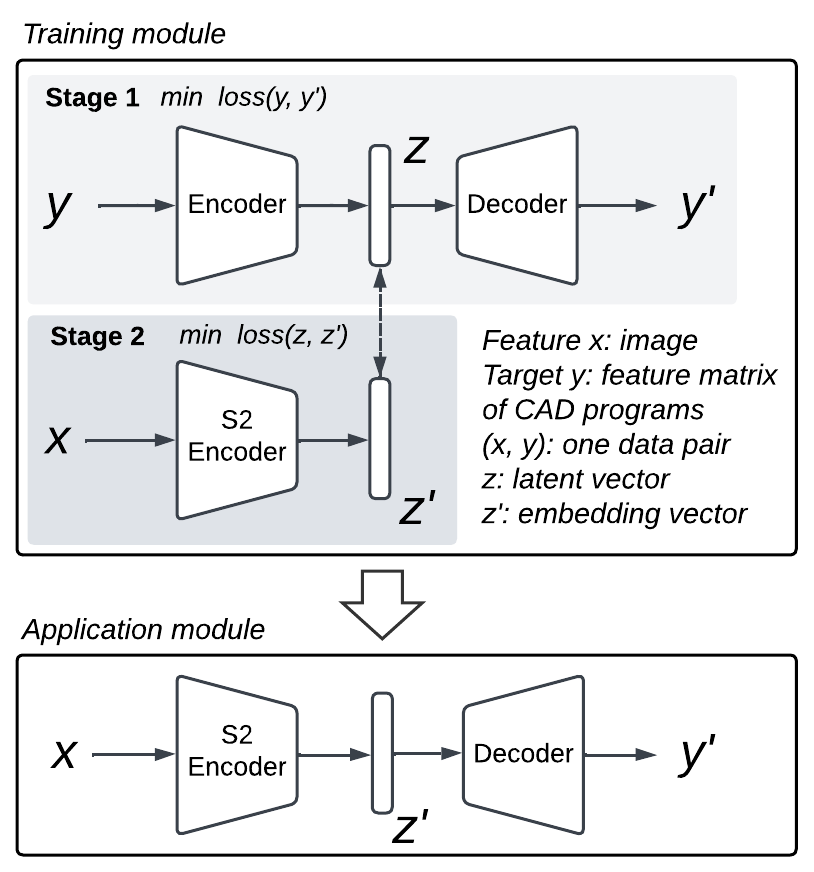}
    \caption{Image2CADSeq model using a target-embedding representation learning method}
    \label{fig:target_embedding}
\end{figure}

\subsection{NEURAL NETWORK MODEL ARCHITECTURE, TRAINING, AND APPLICATION}
\label{sec: network}
The application of target-embedding representation learning in deep learning, particularly for cross-modal tasks, has shown considerable efficacy \cite{jarrett2020target, li2023deep}. In alignment with this, we have developed the Image2CADSeq model, utilizing a target-embedding representation learning method, as illustrated in Figure \ref{fig:target_embedding}. It features an encoder-decoder network for Stage 1 (S1), which is geared towards unsupervised learning and enables the efficient encoding of target objects (i.e., matrix feature of CAD programs) within a latent space. An additional encoder is integrated into Stage 2 (S2), focusing on supervised learning to regress the previously learned latent space using feature objects (i.e., images) as input. 

The Image2CADSeq model employs a two-stage training strategy \cite{mostajabi2018regularizing, li2022predictive}. In Stage 1, the focus is on independent training of the encoder-decoder network. The objective is to minimize the reconstruction loss between the actual matrix feature of CAD programs (\(y\)) and its reconstructed equivalent (\(y'\)). Completing this stage involves fixing the learnable parameters of the neural network model and saving the learned model, thereby capturing a latent space of y. Stage 2 shifts the focus to independent training of the S2 encoder by minimizing the discrepancy between the latent vector, derived from the learned latent space, and the embedding vector produced by the S2 encoder using an image as input. Importantly, each image used in this stage is directly associated with its feature matrix from Stage 1. This image and its corresponding feature matrix are associated with the same 3D object, and they form one data pair. 
The alignment of the latent vector with the embedding vector is performed specifically for these data pairs, ensuring that the S2 encoder training is precisely tuned to the corresponding images. This approach ensures a cohesive and targeted learning process. We present a novel data synthesis pipeline to generate training data pairs in Section \ref{sec: data synthesis pipeline}. 

After training the Image2CADSeq model, the S2 encoder is integrated with the decoder from S1, creating the application module. This module is capable of predicting a feature matrix given an image input. Subsequently, this feature matrix can be translated into a CAD program using the Sim-Gallery DSL. Finally, the CAD program can be parsed into a 3D object by utilizing Fusion 360 software. 

\subsection{DATA SYNTHESIS PIPELINE}
\label{sec: data synthesis pipeline}
With the proposed design representation of the CAD programs and Fusion 360 software, we introduce an automatic data synthesis pipeline, as illustrated in Figure \ref{fig:data_synthesis_diagram}. This method is tailored to generate training data pairs comprising feature matrices of CAD programs and the corresponding images, essential for training the Image2CADSeq model.
The process begins with preparing a list of basic shape templates, such as cylinders, employing the \textit{Sketch-and-Extrude} paradigm of the Gallery DSL. For these basic shapes, we establish a series of template operations (e.g., $add\_line, add\_circle$). The corresponding parameter values of these operations are then generated based on the range specified in Table \ref{tab:vec_rep}.
By integrating these template operations with their respective parameters, a complete CAD program is formulated, which is then translated into 3D CAD models through Fusion 360 software. These models are then rendered to obtain their images. Additionally, the CAD programs are vectorized and quantized to derive their feature matrices, as discussed in Section \ref{sec: vector rep}. An image paired with its feature matrix, both derived from the same CAD program, constitutes a data pair. The method is exemplified using a cylinder model in Figure~\ref{fig:data_synthesis_diagram}.

\begin{table*}
    \caption{Comprehensive evaluation metrics for the image, the CAD program, and the 3D CAD model}
    \label{tab:metrics}
    \centering
    \includegraphics[width=0.8\textwidth]{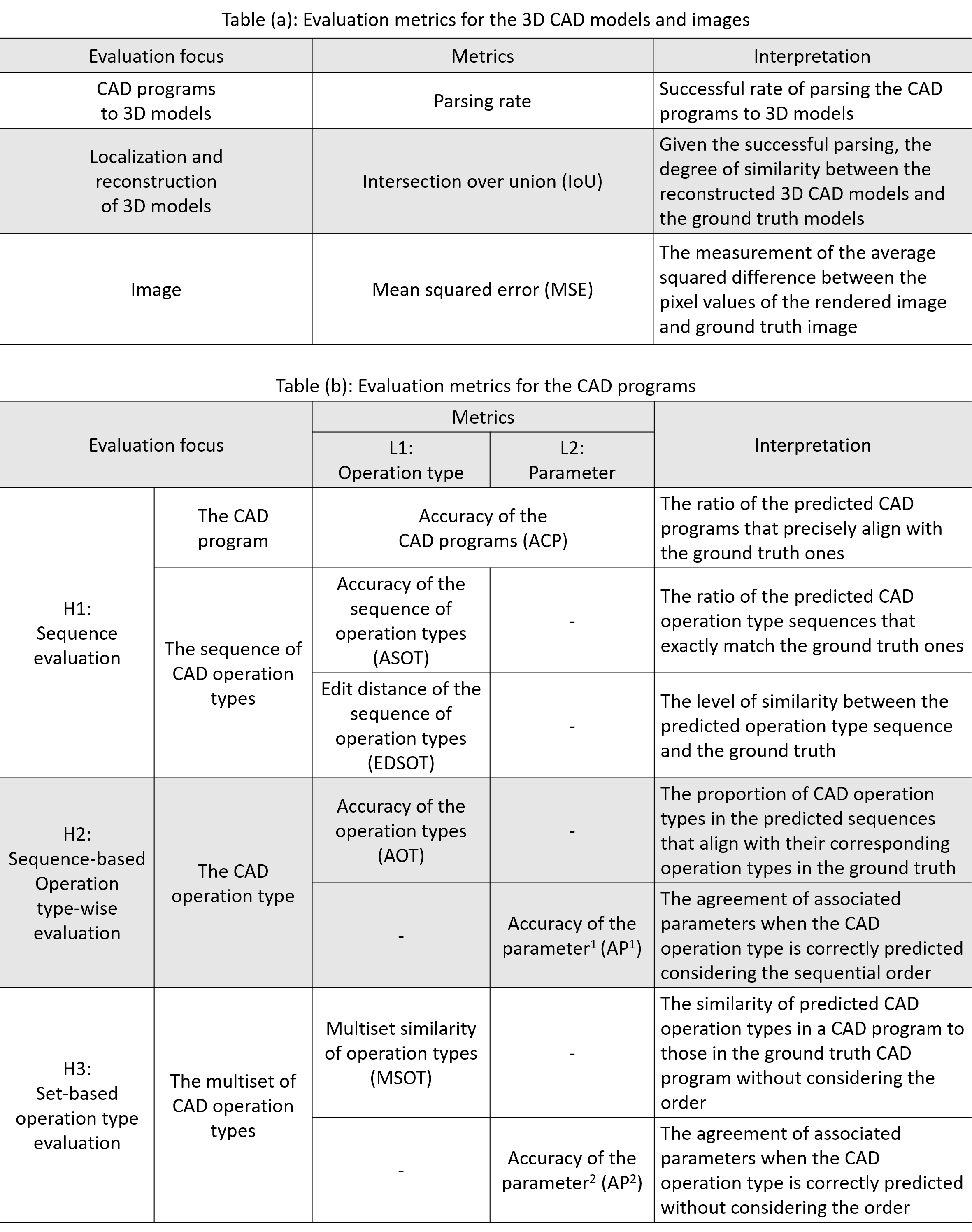}
\end{table*}

\subsection{EVALUATION METRICS}
\label{sec:metrics}
In the Image2CADSeq task, there are three key elements: the image, the CAD program, and the 3D CAD model. The CAD program consists of a sequence of CAD operations, each involving an operation type and its associated parameters. To assess the effectiveness of our approach, we have developed a set of evaluation metrics, as illustrated in Table~\ref{tab:metrics}.

For the evaluation of 3D CAD models and images, we utilize established metrics such as the intersection over union (IoU) and mean squared error (MSE), respectively, as shown in Table \ref{tab:metrics}(a). However, assessing the quality of CAD programs poses a challenge due to the scarcity of suitable metrics in the literature. To address this gap, we introduce a novel evaluation system for CAD programs based on the proposed matrix representation, detailed in Table \ref{tab:metrics}(b). 
To comprehensively evaluate the information loss between the predicted CAD programs with the ground truth, this system incorporates both hierarchical (H1-3) and double-layered (L1,2) aspects as shown below, facilitating a multi-dimensional assessment of CAD program prediction.  
We expect this evaluation system to become a standard for tasks involving the prediction of CAD programs, as further discussed in the following.

\begin{enumerate}[label=\arabic*.]
  \item Hierarchies:
  \begin{itemize}[label=\textbullet, left=0em]
    \item H1: Sequence evaluation -- Evaluates the accuracy of the entire CAD program and the specific order in which certain CAD operation types follow.
    \item H2: Sequence-based operation type evaluation -- Examines the accuracy of each individual operation type within the sequence.
    \item H3: Set-based operation type evaluation -- Assesses the operation types as a collective set, without considering the sequential order. Even if the operation type sequence varies, a prediction is considered superior if it accurately predicts a higher number of operation types due to the preservation of information.
  \end{itemize}
  \item Layers:
  \begin{itemize}[label=\textbullet, left=0em]
    \item L1: Operation type layer -- Evaluates the accuracy of the CAD operation types.
    \item L2: Parameter layer -- Assesses the accuracy of the parameters associated with each CAD operation type.
  \end{itemize}
\end{enumerate}

Recall that a feature matrix \(P\) can be expressed as $P = \begin{bmatrix} \mathbf{o^1,} & \mathbf{o^2,} & \ldots & \mathbf{, o^{N_c}} \end{bmatrix}^T$. 
The vector \(\mathbf{o^i}\in \mathbb{R}^7\) is a CAD operation vector which can be noted as $ \mathbf{o^i} = \begin{bmatrix} t \\ \mathbf{p} \end{bmatrix} $, 
where \(t\) is an integer indicating the operation type, and \(\mathbf{p}\) is a vector of integers representing the corresponding parameters (see Section \ref{sec:methodology} for more details). In what follows, we explain the evaluation metrics using the same notation. 

\textbf{ACP.} Accuracy of CAD programs (ACP) is calculated by Equation \eqref{eq:ACP}, representing the ratio of the predicted CAD programs that are precisely aligned with the ground truth ones, where $N$ is the total number of test data for evaluation, $P^i$ denotes the ground truth CAD program, while $\hat{P^i}$ represents the corresponding predicted CAD program, $\mathbb{I}(\cdot)$ is the indicator function that returns 1 if the condition is true, and 0 otherwise.

\begin{equation}
\label{eq:ACP}
    \mathrm{ACP}=\frac{1}{N}\sum_{i=1}^{N}\mathbb{I}(P^i=\widehat{P^i}),
\end{equation}

\textbf{ASOT \& EDSOT.} Two metrics are defined for evaluating the sequence of operation types: 1) accuracy of the sequence of operation types (ASOT) and 2) edit distance of the sequence of operation types (EDSOT). ASOT assesses the proportion of predicted CAD sequences with operation types (without considering the associated operation parameters) that match exactly the ground truth, as defined by Equation \eqref{eq:ASC}. The notation adheres to those introduced in Equation \eqref{eq:ACP}. In addition, $P^i[:,1]$ represents the first column of $P^i$ which is the sequence of operation types in a CAD program, and similarly for $\hat{P^i}[:,1]$.

\begin{equation}
\label{eq:ASC}
    \mathrm{ASOT}=\frac{1}{N}\sum_{i=1}^{N}\mathbb{I}(P^i[:,1]=\widehat{P^i}[:,1]),
\end{equation}

\begin{equation}
\label{eq:EDSC}
M[i,j] = \begin{cases}
\max(i, j) & \text{if} \min(i,j)=0, \\
\min \begin{cases}
M[i-1,j] + 1 \\
M[i,j-1] + 1 \\
M[i-1,j-1] + 1_{(a_i \neq b_j)}
\end{cases} & \text{otherwise.}
\end{cases}
\end{equation}

In the case of EDSOT, it measures the level of similarity between the predicted CAD operation type sequence and the ground truth. While there exist various metrics to calculate the edit distance, we utilize the Levenshtein distance as shown in Equation \eqref{eq:EDSC}, which is commonly employed to compare sequential data in applications, such as computational biology \cite{navarro2001guided}. Given two strings $a$ and $b$ of lengths $m$ and $n$, respectively, the Levenshtein distance $L(a, b)$ can be calculated using dynamic programming. We define a matrix $M$ of size $(m+1) \times (n+1)$, where $M[i,j]$ represents the minimum number of operations (i.e., insertions, deletions or substitutions) required to transform the substring $a[1:i]$ into the substring $b[1:j]$.
After calculating the values for all entries of the matrix $M$, the Levenshtein distance is given by $L(a, b) = M[m,n]$.

\textbf{AOT.} As shown in Equation \eqref{eq:AC}, the accuracy of the operation types, denoted as AOT, is computed as the proportion of CAD operation types in the predicted sequences that align with their corresponding operation types in the ground truth, taking into account the order. 
Common notations are used as in previous equations. In addition, the function $|\cdot|$ is employed to determine the length of a sequence, and $l^i$ is defined as $\min (|P^i[:,1]|, |\widehat{P^i}[:,1]|)$, representing the number of operation types that need to be compared in a sequence for the $i$th data point of the test data for evaluation. 
\begin{equation}
\label{eq:AC}
    \mathrm{AOT}=\frac{\sum_{i=1}^{N}\sum_{j=1}^{l^i}\mathbb{I}(P^i[j, 1]=\widehat{P^i}[j,1])}{\sum_{i=1}^{N}|(P^i[:,1])|}
\end{equation}

\textbf{AP}$^1.$ The accuracy of parameter$^1$ (AP$^1$) is determined by assessing the agreement of associated parameters when the CAD operation type is correctly predicted considering the sequential order, as defined in Equation \eqref{eq:ap1}. Conditions (c1-3) serve as the input criteria for the indicator functions. This metric function serves as the second layer beneath the first layer, AOT, indicating that parameter evaluation occurs exclusively when the operation type is accurately predicted (i.e., c1). For c2, recall our use of 8-bit integers (i.e., $0-255$) to represent the parameter values. Regarding c3, the parameter $\eta$ denotes the permissible tolerance for differences between the predicted parameters and their ground truth values. For example, given a specific permissive tolerance $\eta \in[0,255]$, if a ground truth parameter value is $z\in[0,255]$, to be counted as a correct prediction, the predicted parameter value $\hat{z}$ must satisfy the following conditions: $|\hat{z}-z|\leq \eta$ and $\hat{z}\in[0,255]$. 
Furthermore, the summation over $k$ is a consequence of each CAD operation vector $\mathbf{o} \in \mathbb{R}^7$ having its first dimension representing the operation type, while the subsequent dimensions (i.e., dimensions 2-7) pertain to the associated parameters.

\begin{equation}
\label{eq:ap1}
\fontsize{10pt}{12pt}\selectfont
\begin{aligned}
    \mathrm{AP^1} &= \frac{\sum_{i=1}^{N}\sum_{j=1}^{l^i}\sum_{k=2}^{7}(\mathbb{I}(\mathrm{c1}) \cdot\mathbb{I}(\mathrm{c2})\cdot\mathbb{I}(\mathrm{c3}))}{\sum_{k=2}^{7}\sum_{i=1}^{N}|(P^i[:,1])|}\\
    \mathrm{c1} &: P^i[j, 1]=\widehat{P^i}[j,1] \\
    \mathrm{c2} &: 0\leq \widehat{P^i}[j,k]\leq 255 \\
    \mathrm{c3} &: |P^i[j, k]-\widehat{P^i}[j,k]|\leq\eta
\end{aligned}
\end{equation}

\textbf{MSOT.} The multiset similarity of operation types (MSOT) is a metric that compares the similarity of predicted CAD operation types in a CAD program to those in the ground truth CAD program, without taking the sequential order into account. In mathematics, a set is defined as a collection of elements where the order of these elements is irrelevant and duplicate elements are not permitted. Conversely, a multiset follows a similar principle as a set, but it allows the inclusion of repeated elements. Thus, a set can be seen as a special case of multiset where each element occurs only once.
To implement the MSOT, we adapted two commonly used metrics: Tanimoto coefficient (TC) and cosine similarity (CS) in the Cheminformatics field for carrying out molecular similarity calculations \cite{bajusz2015tanimoto}.
As the order of the elements in a multiset is not concerned in this case, we can represent a multiset as a vector, where each element corresponds to the count of a particular element in the multiset. For instance, in this study, we have a universe of elements for all the operation types $\{0, 1, 2, 3, 4, 5, 6\}$, a multiset of a triangular prism  $\{5, 0, 1, 1, 1, 4, 6\}$ can be represented by a vector $[1, 3, 0, 0, 1, 1, 1]$ with each number representing the count of a particular operation type. Denote $\mathbf{a}$ and $\mathbf{b}$ as the vectors of two multisets and the TC between $\mathbf{a}$ and $\mathbf{b}$ can then be calculated as Equation \eqref{eq:tc}, where $\mathbf{a} \cdot \mathbf{b}$ denotes the dot product between the two vectors (sum of the element-wise multiplication), $||\cdot||$ denotes the Euclidean norm. CS measures the cosine of the angle between two vectors and CS between $\mathbf{a}$ and $\mathbf{b}$ is calculated as Equation \eqref{eq:cs}. 
\begin{equation}
\label{eq:tc}
    \mathrm{TC}(\mathbf{a}, \mathbf{b}) = (\mathbf{a}\cdot \mathbf{b}) / (||\mathbf{a}||^2 + ||\mathbf{b}||^2 - \mathbf{a}\cdot \mathbf{b})
\end{equation}

\begin{equation}
\label{eq:cs}
    \mathrm{CS}(\mathbf{a}, \mathbf{b}) = (\mathbf{a} \cdot \mathbf{b}) / (||\mathbf{a}|| \times ||\mathbf{b}||)
\end{equation}

\textbf{AP}$^2.$ Similar to AP$^1$, the accuracy of parameter$^2$ (AP$^2$) serves as the second layer in the evaluation of CAD operations which can be similarly calculated using Equation \eqref{eq:ap1}. 
Notably, in AP$^2$, the assessment does not take into account the order of operations. In this study, an operation type can occur multiple times in a CAD program, such as the \textit{Line} operation in a triangular prism. This introduces a challenge regarding which instance of the \textit{Line} operation in the predicted CAD program should be matched with the corresponding instance in the ground truth CAD program for parameter comparison. Despite this challenge, AP$^2$ retains practical significance, particularly in scenarios where there are no repeated elements in the CAD operations. In addition, it is essential to maintain AP$^2$ to preserve the integrity of the evaluation system.

\section{EXPERIMENTS AND RESULTS}
\label{IMPLEMENTATION DETAILS AND RESULTS}
In this section, we introduce our experiments and results in detail. We begin with the introduction to training data preparation, followed by our experiments on different strategies of synthesizing datasets, different neural network architectures, and then, the results. 

\begin{table}
  \caption{Template shapes for the synthesis of training data}
  \label{tab:template_shapes}
  \centering
  \includegraphics[width=0.8\linewidth]{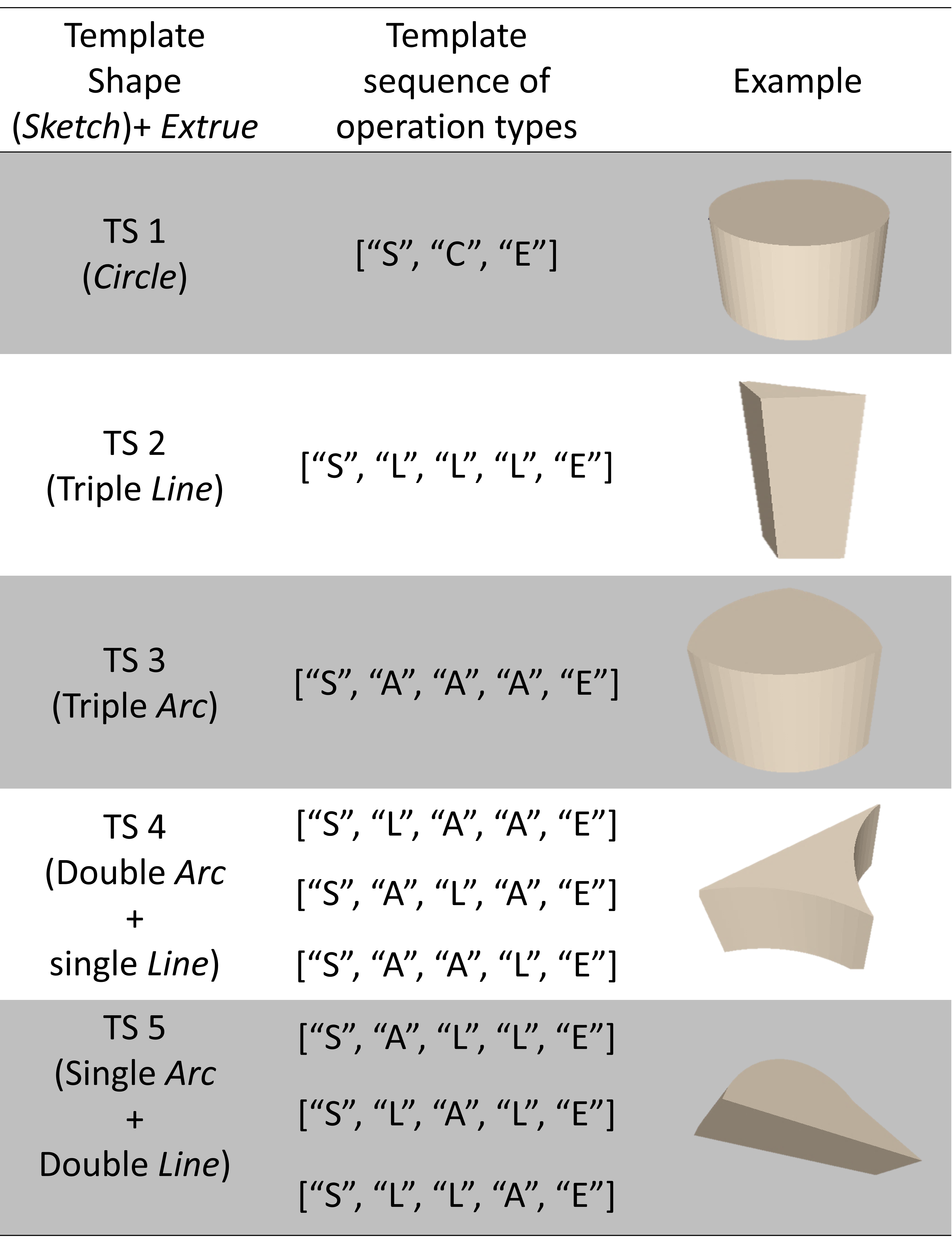}
\end{table}

\subsection{TRAINING DATA PREPARATION}
Based on the data synthesis pipeline outlined in Section \ref{sec: data synthesis pipeline}, we developed a collection of 5 template shapes (TS), as depicted in Table \ref{tab:template_shapes}. These shapes are crafted using \textit{Sketch-and-Extrude} operations, detailed in Table \ref{tab:gallery_dsl}. To create the \textit{Sketch} of each shape, we utilized \textit{Line}, \textit{Arc}, and \textit{Circle} operations, and applied the \textit{Extrude} operation to generate the 3D volume of these shapes. TS 1-3 each correspond to unique sequences of operation types, while TS 4 and 5 are associated with three varied sequences. An example for each template shape is also presented. 

We employ two distinct strategies for dataset preparation:
(1) Random generation of CAD programs (i.e., dataset without rules): We assign random values to CAD operation parameters for the template sequences. This randomness is within predefined value ranges as outlined in Table \ref{tab:vec_rep}. For example, for an $add\_line(\cdot)$ operation, we randomly select a number from the range $[-1, 1]$ to determine the coordinates of endpoints. This method ensures diversity in the dataset by incorporating a wide range of parameter values, reflecting various potential CAD designs.
(2) Generation of CAD programs with embedded design rules (i.e., dataset with rules): Contrary to the random approach, this method involves parameter selection based on the design rules that we defined. For example, the extrusion depth of a circle is determined by the coordinates of its center point. This strategy mimics the purpose-driven process typical of real-world design scenarios. 
By incorporating design rules into data synthesis, we embed design knowledge in the data. This method is expected to test how the embedded design principles would affect the effectiveness of CAD reconstruction from images. The combination of random generation and rule-based design in dataset preparation allows for a comprehensive evaluation of our system's capabilities in varying scenarios. 

Under each synthesis strategy, we synthesized 2,000 shapes corresponding to every sequence type outlined in Table \ref{tab:template_shapes}, except for the sequence of TS 1, for which we synthesized 6,000 shapes. This was taken to ensure a balanced dataset in terms of both the length of sequences and the number of shapes for each template shape category. Consequently, this led to the creation of 22,000 CAD models for each strategy. These models were derived by processing the synthesized CAD programs through Fusion 360 software. For imaging purposes, all these 3D models were rendered using a uniform perspective camera positioned at $(20.0, 20.0, 20.0)$ looking towards the origin $(0.0, 0.0, 0.0)$ and all the images are in the resolution of $512\times512$ pixels. This process resulted in the image set $X=\{x_k\}_{k=1}^{22000}$. The corresponding feature matrices $Y=\{y_k\}_{k=1}^{22000}$ were also saved during synthesis. The final training dataset was $\{x_k, y_k\}_{k=1}^{22000}$.

\begin{figure}
    \centering
    \includegraphics[width=1.0\linewidth]{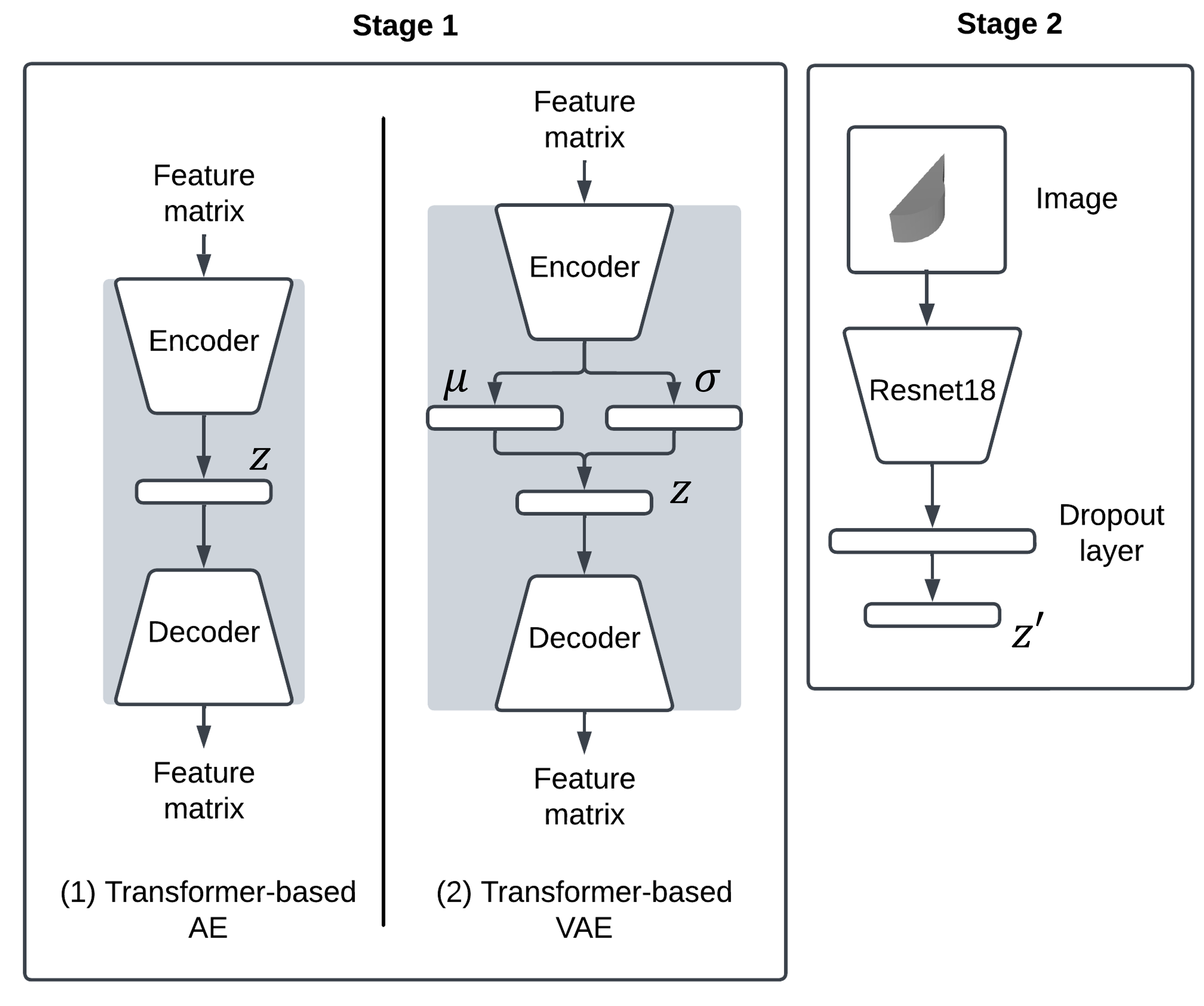}
    \caption{Implementation of the Image2CADSeq model. Two different encoder-decoder architectures were explored in Stage 1: (1) Baseline model: a transformer-based autoencoder (AE), adapted from DeepCADNet \cite{wu2021deepcad}, and (2) Enhanced model: a transformer-based variational autoencoder (VAE), which extends the AE architecture. In Stage 2, the encoder is developed based on ResNet18 \cite{he2016deep}, employing a dropout layer before the final layer to mitigate overfitting and enhance generalization.}
    \label{fig:implement_image2cadseq_pro}
\end{figure}

\subsection{IMAGE2CADSEQ MODEL ARCHITECTURES AND TRAINING}
In Figure \ref{fig:implement_image2cadseq_pro}, we illustrate the development of the Image2CADSeq model. 
The construction of the model involved the application of target representation learning techniques, commonly employing target-embedding autoencoders (TEA) that utilize an autoencoder to obtain the latent representation of target objects \cite{jarrett2020target}. In a recent development, Li et al. \cite{li2022predictive} introduced a target-embedding variational autoencoder (TEVAE) by extending the autoencoder to a variational autoencoder (VAE). This approach has demonstrated superior effectiveness in the cross-modal synthesis of 3D designs. However, their study did not include an empirical comparison between the two architectures in terms of the generated results. Therefore, as part of our study's contributions, we investigated two models: 1) a baseline transformer-based AE and 2) an improved version in the form of a transformer-based VAE for Stage 1. The aim was to compare their performance and explore a better architecture for the Image2CADSeq model.

We modified the first and last layer of the transformer-based AE, adapted from DeepCAD \cite{wu2021deepcad}, to capture and reconstruct the feature matrices of the CAD programs in this study. Its primary objective is to learn and interpret the latent space of these matrices, providing a robust foundation for accurate feature representation. The AE model is used to establish a baseline for understanding and processing the complex structures inherent in CAD designs. It uses a typical reconstruction loss coupled with a regularization loss. Reconstruction loss ensures accurate reconstruction of input features in the output, while regularization loss prevents overfitting, promoting a more generalized model capable of handling various CAD designs.

Extending the AE architecture, the VAE introduces a probabilistic approach to encoding, which is tailored to construct a smoother latent space, surpassing the AE in terms of flexibility and adaptability. 
The VAE employs a more complex loss function with KL-divergence loss, reconstruction loss, and regularization loss. The KL-divergence loss is pivotal in managing the probabilistic aspect of VAE, ensuring that the encoded distributions are effectively regularized. This, along with the reconstruction and regularization losses, forms a comprehensive approach to learning, capturing both the variance and the intricate details of CAD sequences. Stage 2 of the development incorporates an encoder based on ResNet18 \cite{he2016deep}. In addition, a dropout layer is positioned between the encoder and the embedding vector layer to prevent overfitting to the training data, thus maintaining its efficacy on unseen data. We utilized a regression loss between the embedding vector of the image and its corresponding latent vector obtained from the latent space in Stage 1 and a regularization loss to promote the generalizability of the Stage 2 encoder. 

We divided each of the two training datasets, i.e., the dataset with or without rules, into three subsets: train, validation, and test set with a proportion of 8:1:1. The validation set was used to monitor the training process, preventing the model from overfitting to the train set data and ensuring that the model's generalization capabilities to the unseen data, e.g., test set data. 
We employed a grid search strategy in Stage 1 to find optimal hyperparameters of the neural network models and the training, aiming to minimize the training loss while maintaining good generalizability of the models. The AE and VAE models exhibited similar training trends, leading us to use the same hyperparameter set for both models. 
In our experiments, for Stage 1, a latent dimension size of 256 proved optimal for both models, resulting in the lowest reconstruction loss for the test set data among trials with dimensions of 64, 128, 256, and 512.
Other hyperparameters include 500 epochs of training with a batch size of 512, the Adam optimizer, and a learning rate of 0.001. 
Moving to Stage 2, we initiated training with a pre-trained ResNet18 model \cite{he2016deep} that possesses a broad comprehension of various images.
The S2 encoder was trained for 50 epochs using the Adam optimizer with a learning rate of 0.0001 and a batch size of 128. A dropout ratio of 0.4 was applied in the dropout layer.

\begin{figure*}
    \centering
    \includegraphics[width=\textwidth]{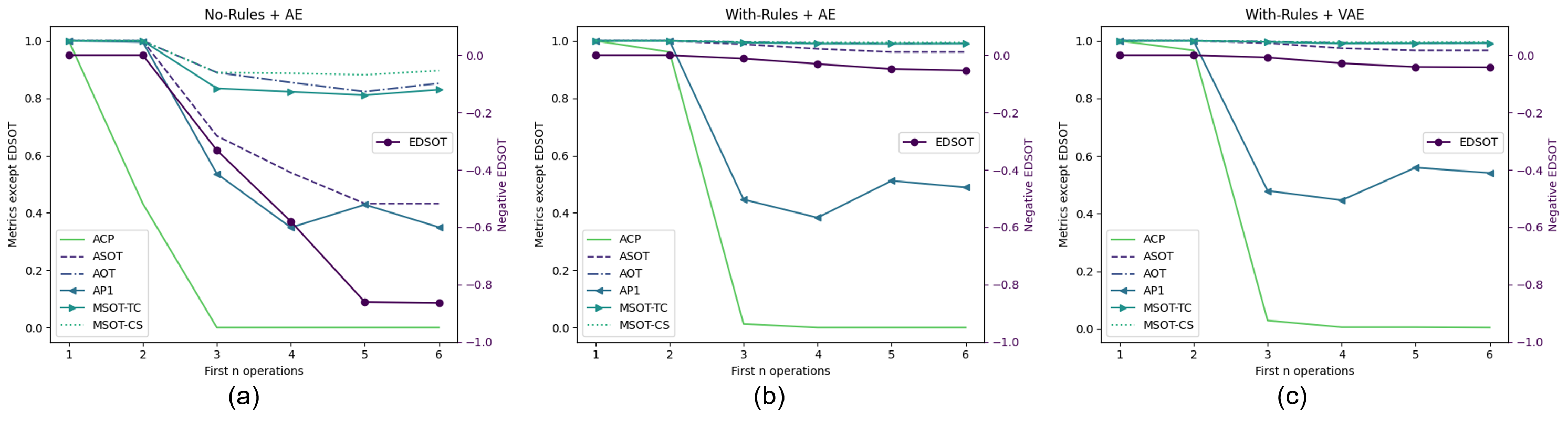}
    \caption{Evaluation of the Image2CADSeq model's performance using two distinct architectures with two datasets. (a) Case 1: Results from the network utilizing the TEA architecture trained on the dataset without rules. (b) Case 2: Results from the same TEA architecture but trained on the dataset with rules. (c) Case 3: Results using the TEVAE architecture trained with the dataset with rules. Each figure illustrates the variation of the network's performance metrics (shown in Table \ref{tab:metrics}) versus the first $n$ CAD operations in a CAD program. Specifically, a tolerance $\eta=3$ is chosen for metrics that involve the calculation of the accuracy of parameters, such as ACP and AP$^1$.}
    \label{fig:results_multi_metrics}
\end{figure*}

\subsection{RESULTS}
In this section, we present the results of our experiments on the performance of the Image2CADSeq model.
\label{subsec:results}
\subsubsection{OVERALL EVALUATION OF THE CAD PROGRAMS}
Figure \ref{fig:results_multi_metrics} provides a comparison of the Image2CADSeq model's performance, evaluated under two different architectures and two datasets. Specifically, there are three subfigures, each representing a unique combination of architecture and dataset. Figure \ref{fig:results_multi_metrics} (a) illustrates the performance of the network when employing the TEA architecture in conjunction with the dataset without rules. 
In contrast, Figure \ref{fig:results_multi_metrics} (b) presents results derived from the same TEA architecture, but the network is trained on the dataset with rules. We observed a significant improvement when employing the dataset with rules for model training. Consequently, we tested the TEVAE architecture using the dataset with rules only, and the results are presented in Figure \ref{fig:results_multi_metrics} (c).

The figures illustrate the model's performance across various metrics (as defined in Table \ref{tab:metrics} (b)) when applied to the first $n$ operations in a CAD program. We limit $n$ to 6 to encompass the longest template sequences. According to Table \ref{tab:template_shapes}, the maximum length of the template sequences is 5 for CAD operations in addition to a non-CAD operation $SOP$, marking the start of a program. 
Typically, higher metric values indicate superior performance, except for the EDSOT, where lower values are preferable. To maintain a uniform direction of performance across all metrics and enhance the readability of the plotting, we present the EDSOT in its negative form in the figures.
Furthermore, when calculating metrics related to parameter accuracy, such as the accuracy of CAD programs (ACP) and the accuracy of parameter$^1$ (AP$^1$), we introduce a tolerance level ($\eta=3$). This tolerance accounts for permissible deviations in the quantized continuous variables that have 256 levels but does not extend to discrete variables, such as the sketch plane identifier that has only 3 levels (refer to Table \ref{tab:vec_rep}). The tolerance reflects the design problem's criteria, allowing certain margins of error in parameter predictions that can be customized in different scenarios.

The metrics, classified into three hierarchical categories in Section \ref{sec:metrics}, include the accuracy of CAD programs (ACP), the accuracy of the sequence of operation types (ASOT), and the edit distance of the sequence
of operation types (EDSOT) for H1 sequence evaluation; the accuracy of the operation types (AOT) and the accuracy of parameter$^1$ (AP$^1$) for the evaluation of the H2 sequence-based operation type; and the multiset similarity of operation types (MSOT) for the evaluation of the H3 set-based operation type. We analyze these results according to this hierarchical structure. 

Upon analyzing the results of Figure \ref{fig:results_multi_metrics} (a) for the TEA trained using the dataset without rules (referred to as Case 1), we observe a downward trend in all metrics as the sequence length increases. This is intuitive, and predicting longer sequences is inherently more difficult for the model. 
The ACP metric drops to zero at $n=3$, indicating that the model struggles to accurately predict the entire CAD program including both the operation types and parameters even when the sequence is relatively short. Notably, the ACP's decrease to roughly 0.4 at $n=2$ suggests the model's specific difficulty in predicting the sketch plane given the input images, because the second CAD operation—\textit{Sketch}—defines the sketch plane's position. In addition, both ASOT and the negative EDSOT metrics exhibit declines beginning with $n=3$, together with the results of ACP, showing the limited sequence prediction capability of the model when the sequences get longer.

Moreover, despite the model's low values in H1 metrics and the low values in AP$^1$ of H2, it scores highly on the AOT metric of H2 and maintains high values of MSOT-TC and MSOT-CS in H3. 
This discrepancy and inconsistency probably result from the characteristics of the dataset, in which different shape categories share similar CAD sequences (see Figure \ref{tab:template_shapes}). For example, a ground truth (GT) sequence of TS 3, $[``S”, ``A”, ``A”, ``A”, ``E”]$, could be mistakenly predicted as a TS 4 sequence, e.g., $[``S”, ``L”, ``A”, ``A”, ``E”]$ by the model. Although such a prediction would be deemed as an incorrect sequence prediction when evaluated against ASOT, it would score well (i.e., 4 correct and 1 incorrect operation type) in terms of AOT due to the correctly predicted operation types. 
A dataset encompassing a broader array of design objects with diverse sequences of CAD operations might mitigate such discrepancies in the metric values, such as real-world designs collected from human designers. More significantly, it underscores the need for a comprehensive evaluation framework for image-to-CAD sequence prediction, ensuring that models are thoroughly assessed from multiple perspectives. Otherwise, the results of the performance of the models might be biased.

\begin{table}
    \caption{Results when evaluated at the first 6 operations of the CAD programs}
    \label{tab:results_first6}
    \centering
    \includegraphics[width=0.7\linewidth]{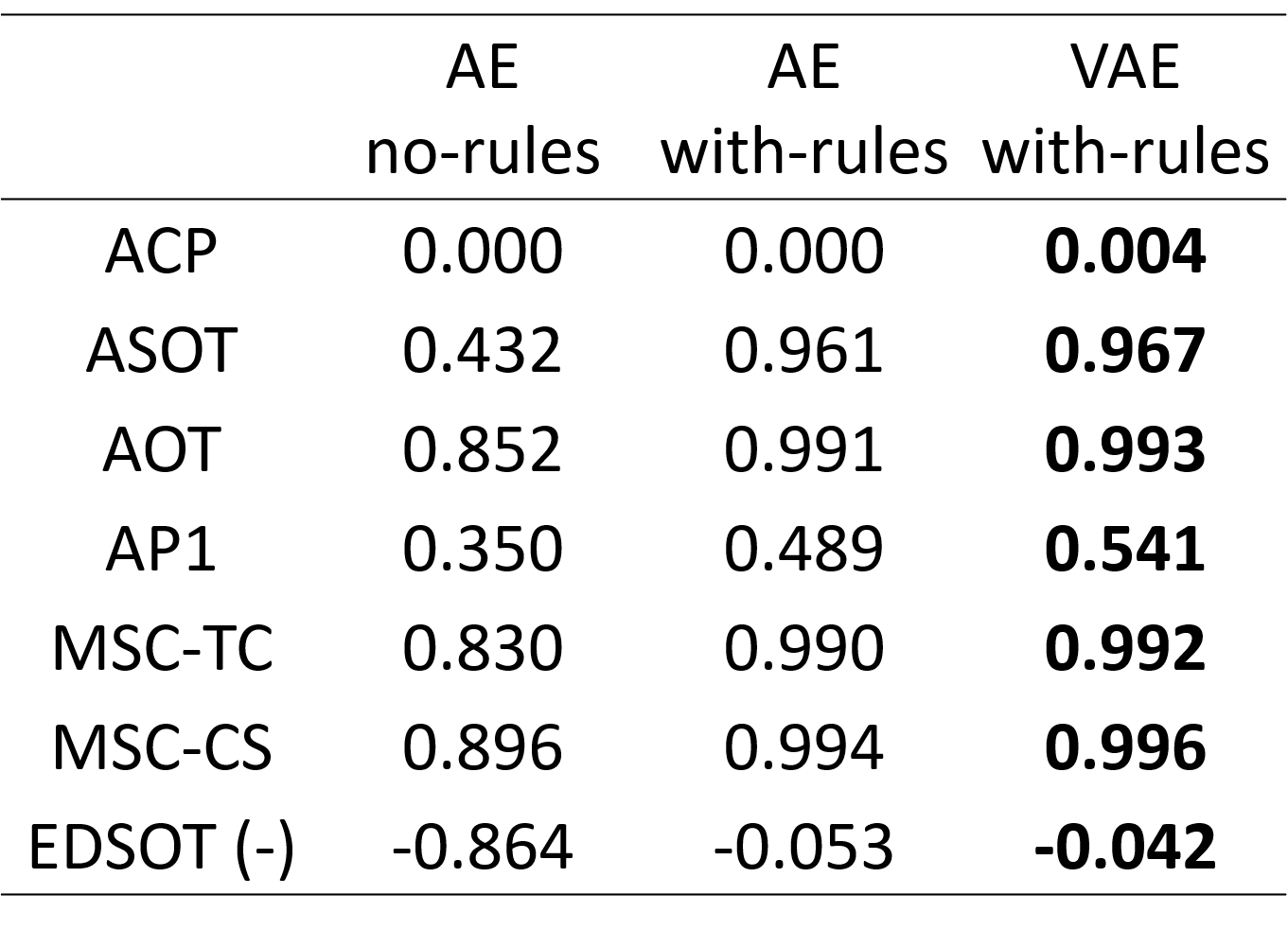}
\end{table}

In Figure \ref{fig:results_multi_metrics} (b), we show the performance of the TEA architecture trained on the dataset with rules (referred to as Case 2), in contrast to the earlier results in Figure \ref{fig:results_multi_metrics} (a). In particular, this treatment improves significantly in most metrics, except for ACP and AP$^1$. These exceptions, however, do not overshadow the overall enhancement in the model's ability to predict sequences accurately. 
Specifically, while ACP does not show a significant improvement, its slower rate of decline at \(n=2\) and 3 indicates an improvement in the model's performance of predicting sketch planes. 
Furthermore, although AP$^1$ does not show a significant improvement, it convergences to a higher value than the previous data treatment, suggesting an improved performance in parameter prediction.

The significant differences between Figures \ref{fig:results_multi_metrics} (a) and (b) highlight the positive impact that design rules can have on the performance of the model in Image2CADSeq predictions. 
However, despite the improvement when including design rules, the model still faces challenges in accurately predicting parameters.

Building upon the insights from the first two cases, we evolved our model from TEA to TEVAE and trained it using the dataset with rules (referred to as Case 3). The results of the TEVAE model are detailed in Figure \ref{fig:results_multi_metrics} (c). It displays high accuracy in most metrics similar to the baseline performance of the TEA model from Case 2 but largely surpasses its performance in ACP and AP$^1$. 
Particularly, the ACP metric shows a significant improvement in the TEVAE model and achieves a higher value at \(n=3\) (does not decrease to zero as in Case 2). The AP$^1$ metric also reveals an upward trend, settling at a higher value than previously seen with the TEA model.
For a more complete comparison of the three cases, we summarize the results of all metrics at $n=6$ in Table \ref{tab:results_first6}. This summary demonstrates that the TEVAE model, when trained using the dataset with design rules, not only surpasses the TEA counterpart using the same dataset but also gives the best results across all metrics evaluated in all three cases. 

\begin{figure}
    \centering
    \includegraphics[width=0.8\linewidth]{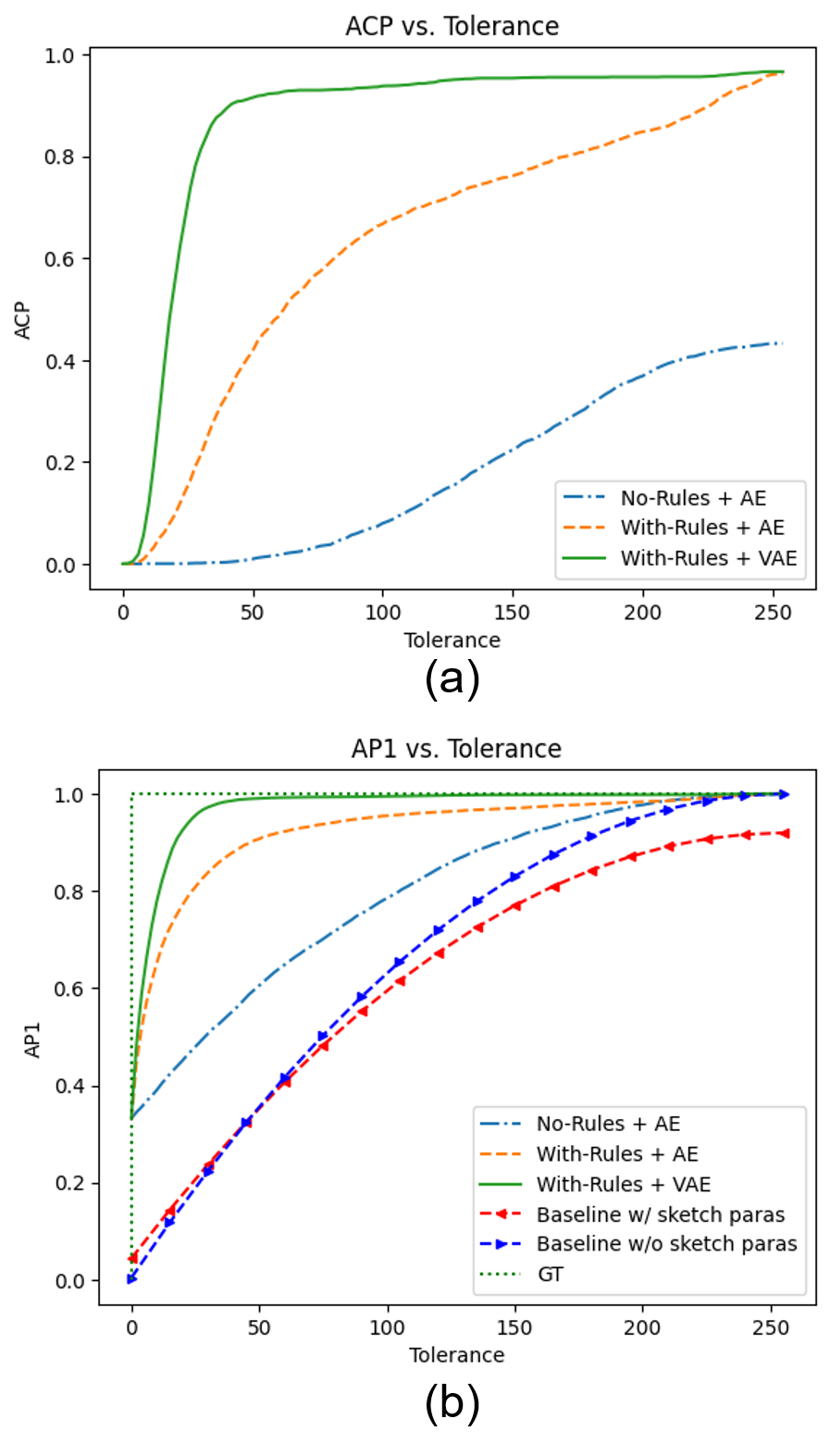}
    \caption{Overall parameter accuracy versus the tolerance levels evaluated using (a) ACP and (b) AP1 for the three cases. Especially, for (b), we included the baseline of the random prediction with or without considering the \textit{Sketch} parameter $I$ and the ground truth line.}
    \label{fig:results_ACP_AP1_vs_tol}
\end{figure}

\begin{figure*}
    \centering
    \includegraphics[width=0.95\textwidth]{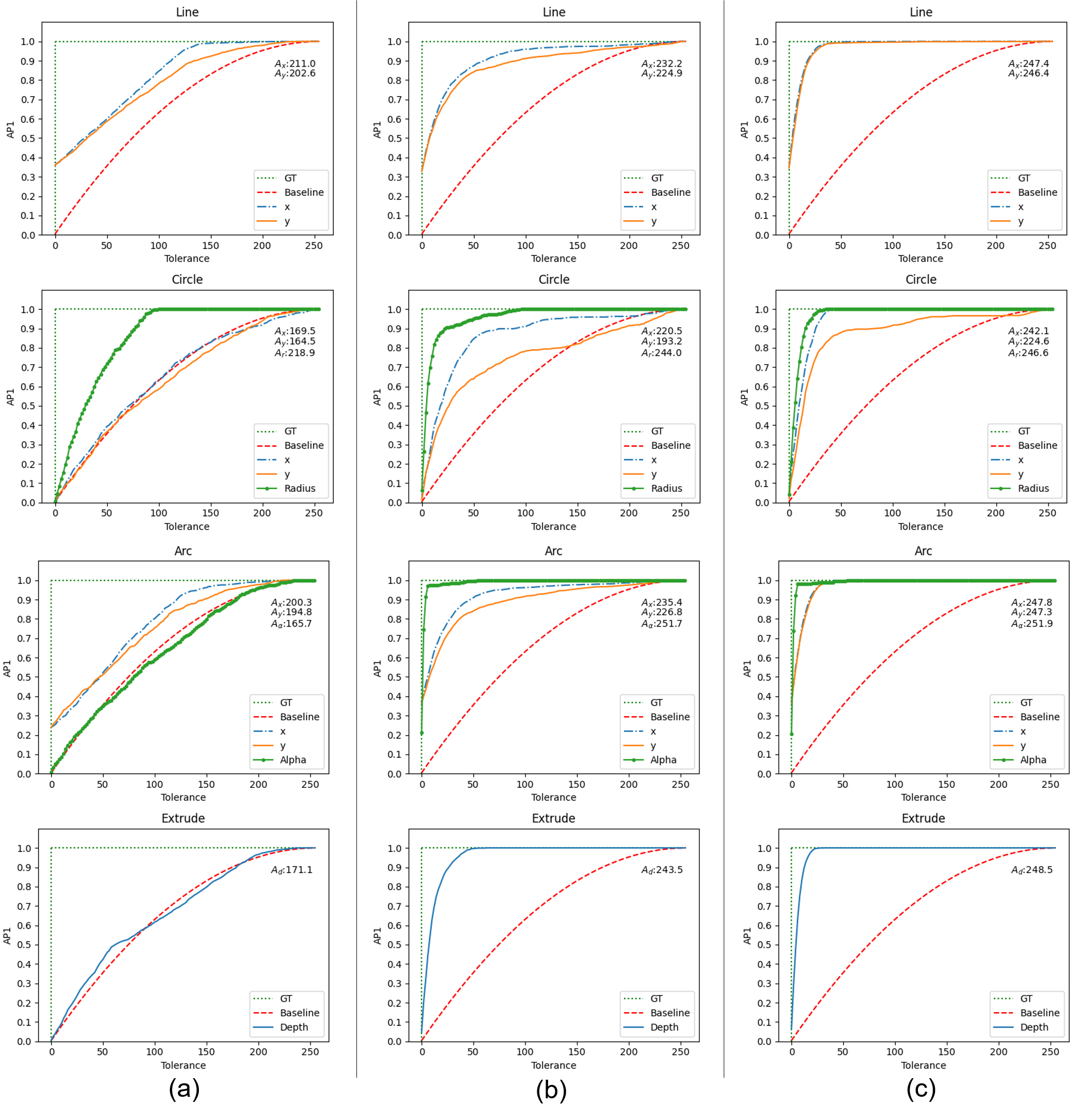}
    \caption{The variation of AP1 versus tolerance for the operation parameters for the \textit{Line}, \textit{Circle}, \textit{Arc}, and \textit{Extrude} operations, in that order. Column (a) shows the results of the network that utilizes the TEA architecture and is trained on the dataset without design rules. Column (b) shows the results of the same TEA architecture but trained on the dataset with rules. Column (c) shows the results using the TEVAE architecture and the dataset with rules.}
    \label{fig:results_AP1_vs_tol_operation_wise}
\end{figure*}

\subsubsection{ANALYSIS ON THE OVERALL PARAMETER ACCURACY}
The models demonstrate high accuracy in predicting the sequence of CAD operations but are less precise in parameter prediction. To facilitate a clearer comparison between the three cases with respect to parameter prediction accuracy, we have included Figure \ref{fig:results_ACP_AP1_vs_tol} to illustrate the relationship between parameter accuracy and tolerance using metrics ACP and AP$^1$. 

Especially, in Figure \ref{fig:results_ACP_AP1_vs_tol} (b), the blue dashed line with triangle markers represents the AP$^1$ value achieved by randomly guessing parameters given a specific tolerance (i.e., the random model), but without considering the \textit{Sketch} parameter (i.e., the identifier of the sketch plane $I$) whose values are not allowed for tolerance. We derived the equation for the random model in Equation \eqref{eq:apwosketch}.
The equation can be simplified to AP$^1=(-\eta^2+511\eta+256)/65536$. This line acts as a baseline to evaluate the model's effectiveness in accurately predicting parameters if the sketch parameter is not considered. 

\begin{equation}
    \label{eq:apwosketch}
    \mathrm{AP}^1=\frac{1}{256}(\frac{(2\eta+1)(256-2\eta)+2(\frac{2\eta(1+2\eta)}{2}-\frac{\eta(1+\eta)}{2})}{256}
\end{equation}

\begin{equation}
    \label{eq:apwithsketch}
    \mathrm{AP}^1=\frac{1}{3} \cdot \frac{11}{91} + \frac{(-\eta^2+511\eta+256)}{65536} \cdot \frac{80}{91}
\end{equation}

To consider the \textit{Sketch} parameter $I$, we need to take into account the characteristics of the dataset (i.e., the ratio of each parameter taken among all possible parameters in the design representation of the CAD programs as outlined in Section \ref{sec: vector rep}). Accordingly, we obtain Equation \eqref{eq:apwithsketch} plotted as the red dashed line with triangle markers in Figure \ref{fig:results_ACP_AP1_vs_tol} (b). Note that the number of the sketch parameter takes $\frac{11}{91}$ of all parameters.
Additionally, a green dotted line is used to indicate the ideal scenario where the parameters are perfectly predicted with zero tolerance (i.e., GT). Other lines in Figures \ref{fig:results_ACP_AP1_vs_tol} (a) and (b) depict the corresponding metric values for different cases, providing a comprehensive view of the model's performance in parameter prediction.

In both (a) and (b) of Figure \ref{fig:results_ACP_AP1_vs_tol}, we consistently see that the metric values increase with rising tolerance levels. A notable point in Figure \ref{fig:results_ACP_AP1_vs_tol} (a) is that the ACP values for all three cases reach their highest at a tolerance of 255 and the corresponding values are 0.432, 0.961, and 0.967 for each case, in accordance with the ASOT values presented in Table \ref{tab:results_first6}. This can be interpreted as the result that when we evaluate the entire CAD program in terms of ACP given that all parameters are accurately predicted, we are essentially assessing ASOT. In Figure \ref{fig:results_ACP_AP1_vs_tol} (b), a crucial observation is that all three lines exceed the baseline of the random guess of parameters. This indicates that the models are effectively learning parameter prediction no matter if there are design rules embedded in the training data or not. 

Additionally, a significant observation in both figures is how differently the models respond to changes in tolerance. Specifically, the TEVAE model, when trained using the dataset with rules, exhibits the highest sensitivity to changes in tolerance in contrast to Cases 1 and 2. This trend suggests that the TEVAE model excels in parameter prediction compared to the TEA model. The accuracy of these predictions depends on both the quality of the training data (for example, in this study, differentiated by the inclusion or exclusion of design rules) and the architecture of the model.

\subsubsection{ANALYSIS OF PARAMETER ACCURACY BASED ON OPERATION TYPES}
To gain more insight into how the models perform in parameter prediction, we plotted the variation of AP$^1$ versus the tolerance for the operation parameters for each CAD operation type.
Spanning columns (a) to (c), the rows in each column show variations in AP$^1$ against tolerance levels (\( \eta = 0-255 \)) for specific parameters, corresponding to different CAD operations, \textit{Line}, \textit{Circle}, \textit{Arc}, and \textit{Extrude}. These results look into the model's adaptability and accuracy across various CAD operations, providing a comprehensive understanding of its capabilities in different model architectures and datasets. Each figure includes a red dashed line representing the baseline as defined in Equation \eqref{eq:apwosketch}. In addition, the green dotted line illustrates the perfect prediction of the parameters with zero tolerance. The other lines show the AP$^1$ for specific parameters related to the respective CAD operations.
To facilitate a more quantitative comparison of how well the parameters are predicted, we also computed the area under the curve (AUC) for each parameter, as indicated in the upper right corner of each figure.

Column (a), the result of Case 1, shows that the $x$, $y$ coordinates of the center of the \textit{Circle}, the sweep angle $\alpha$ of the \textit{Arc}, and the depth $d$ of the \textit{Extrude} align closely with the baseline. This suggests that the TEA model, when trained on a dataset without explicit design rules, performs similarly to random guessing for these specific parameters. However, the model still demonstrates the ability to learn certain patterns from the dataset, as evidenced by its recognition of the endpoint of the \textit{Line}, the radius of the \textit{Circle}, and the endpoint of the \textit{Arc}.  

In Column (b) for Case 2, there is an evident improvement in all metric values compared to Case 1. This improvement highlights the enhanced ability of the TEA model to predict parameters. The significant distance of these values from the baseline indicates that the model has effectively learned the design rules embedded in the training data, enabling it to predict the corresponding parameters more effectively.

In Column (c), the results demonstrate an even better performance. All metric values not only surpass those in Column (b), but they also show a further deviation from the baseline, indicating a significant enhancement of the model's predictive performance. These values are closely approaching the ground truth (GT) line, underscoring the refined ability of the TEVAE model to learn and apply the embedded design rules from the training data. 

\subsubsection{OVERALL EVALUATION OF THE 3D CAD MODELS AND IMAGES}
\begin{figure}
    \centering
    \includegraphics[width=0.9\linewidth]{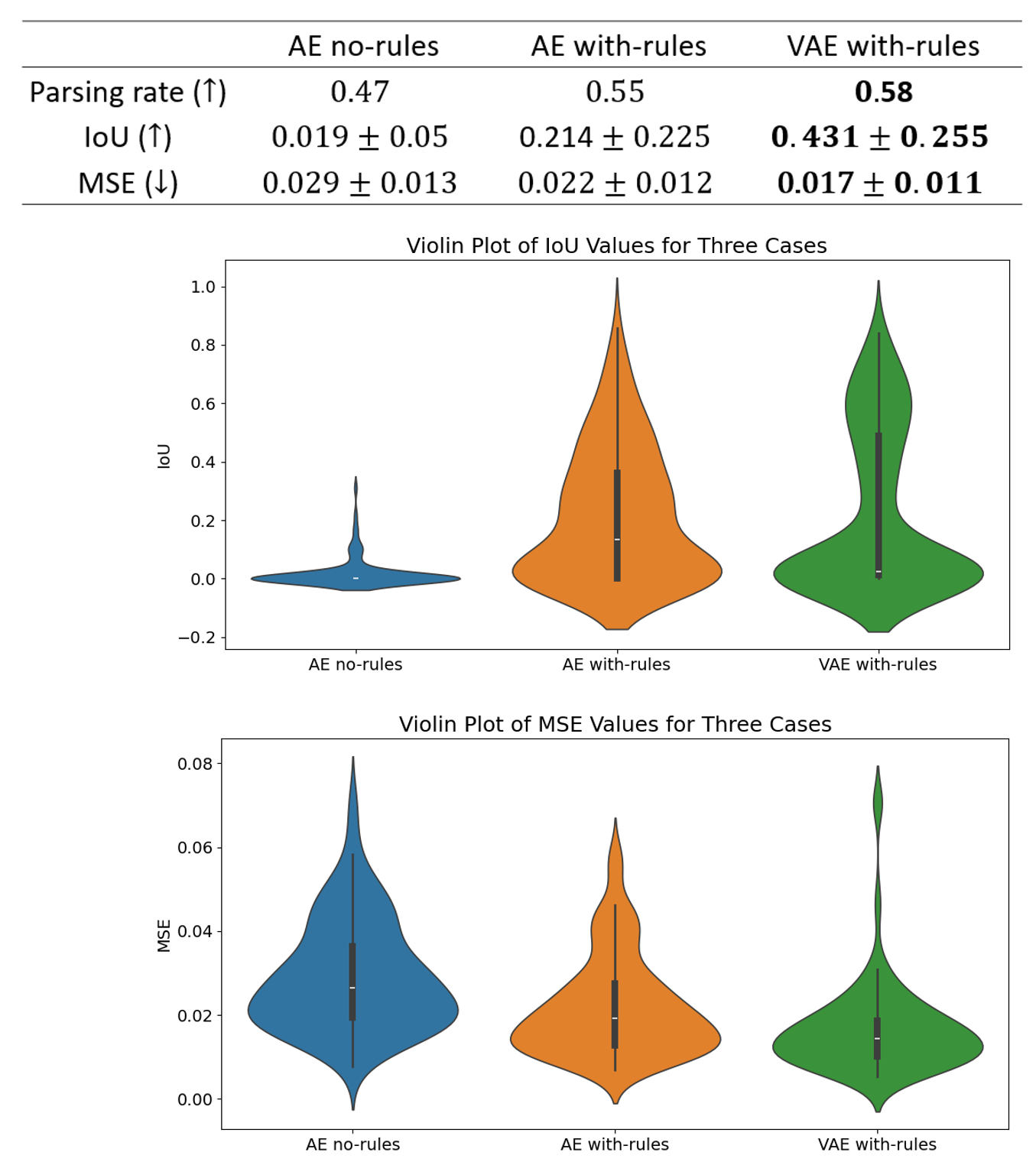}
    \caption{Parsing rate, intersection over union (IoU), and mean squared error (MSE) of the three cases. To analyze IoU and MSE, the mean and standard deviation are computed for each metric within the table. Subsequently, a violin plot is presented illustrating the distribution of these computed values for each metric.}
    \label{fig:3Dandimages}
\end{figure}

Figure \ref{fig:3Dandimages} shows the summary of the parsing rate, intersection over union (IoU), and mean squared error (MSE) outlined in Figure \ref{tab:metrics} of the three cases. We perform an analysis of IoU and MSE by calculating the mean and standard deviation for each metric in the table. 
Following this computation, we depict the distribution of these values for each metric using a violin plot in Figure \ref{fig:3Dandimages}. 
These plots show that the data distributions of the IoU and MSE values move toward improved performance regions (i.e., higher values for IoU and lower values for MSE).
Aligning with our observations in the evaluation of the CAD programs, as introduced in previous sections, the TEVAE trained using the dataset with design rules achieves the best performance among all three cases.

\begin{figure*}
    \centering
    \includegraphics[width=\textwidth]{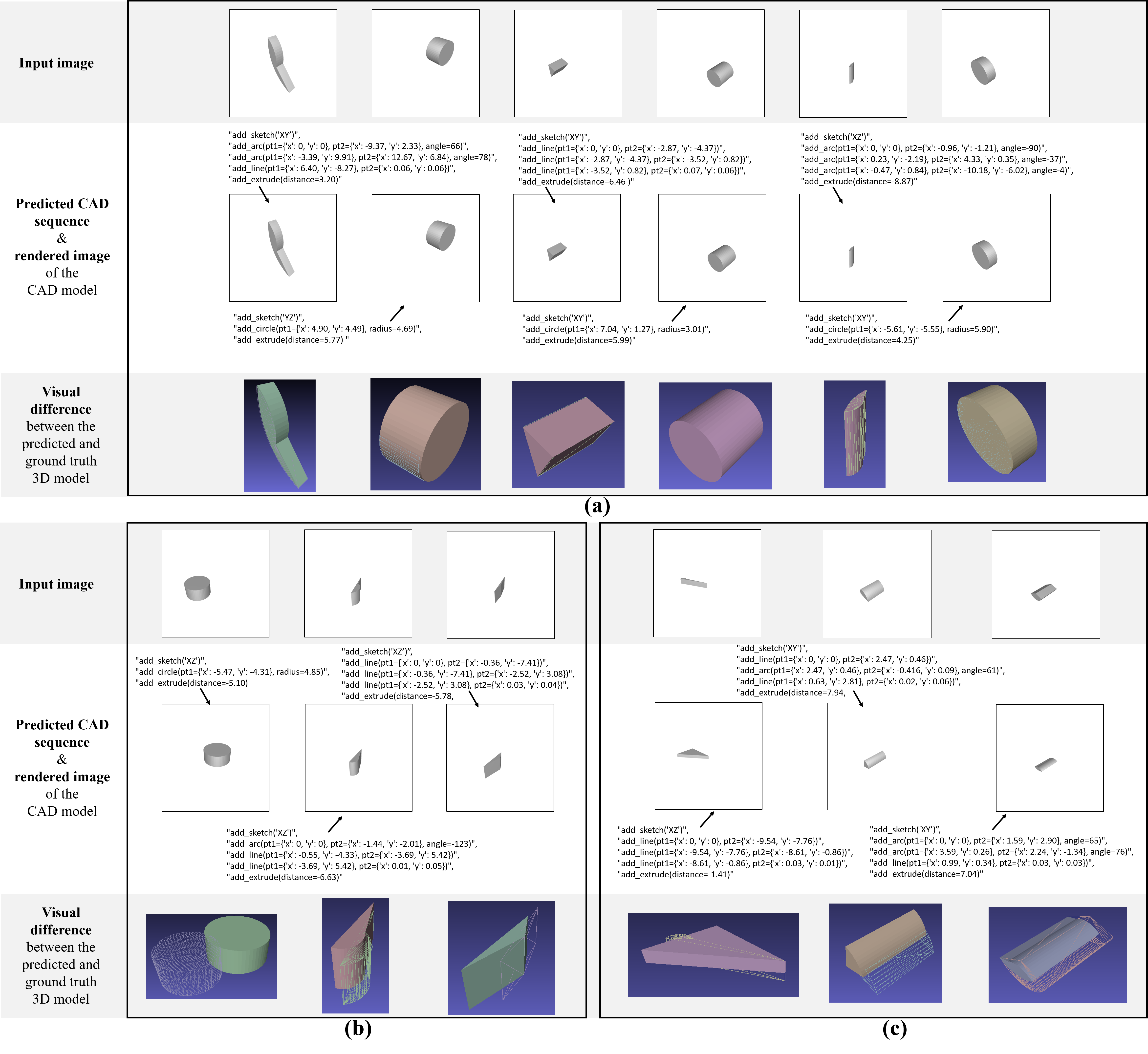}
    \caption{Qualitative analysis of Image2CADSeq model using TEVAE. (a) Near-perfect predictions: High accuracy in shape, size, and position. (b) Satisfactory predictions: Correct shape categories, and inaccurate sizes or positions. (c) Inadequate predictions: Wrong prediction of shape categories. Each subfigure includes the first row - Input image; the second row - Predicted CAD sequence and rendered image of the resultant CAD model; the third Row - Visual comparison between the ground truth (wireframe) and predicted 3D model (solid).
}
    \label{fig:visual_results}
\end{figure*}

We show some qualitative results from the Image2CADSeq model with the TEVAE architecture in Figure \ref{fig:visual_results}.
Subfigure (a) demonstrates near-perfect predictions, characterized by a high degree of accuracy in both the shape category determined by the sequence of CAD operation types and the parameters that determine the size and position of the shape. Subfigure (b) shows satisfactory predictions, where the model correctly identifies shape categories, yet exhibits discrepancies in size and position estimation. Subfigure (c) represents inadequate predictions, marked by the model's failure to accurately predict the categories of shapes.
Each subfigure is arranged in a three-row format. The first row presents the initial input image. This is followed by the second row, which showcases two elements: the predicted CAD sequence and the rendered image of the resulting CAD model. The third row offers a comparative visual analysis, where the ground truth 3D shape is illustrated in a wireframe format against the predicted CAD model, rendered in solid. 

The results indicate that the model is capable of generating a CAD sequence to reconstruct a 3D CAD model by accurately capturing and integrating both the spatial positioning and the geometric details reflected in the input image. However, there remains room for improvement, particularly in predicting the correct CAD sequence and especially the associated parameters, as also shown by the aforementioned quantitative results.

\begin{figure*}
    \centering
    \includegraphics[width=0.9\textwidth]{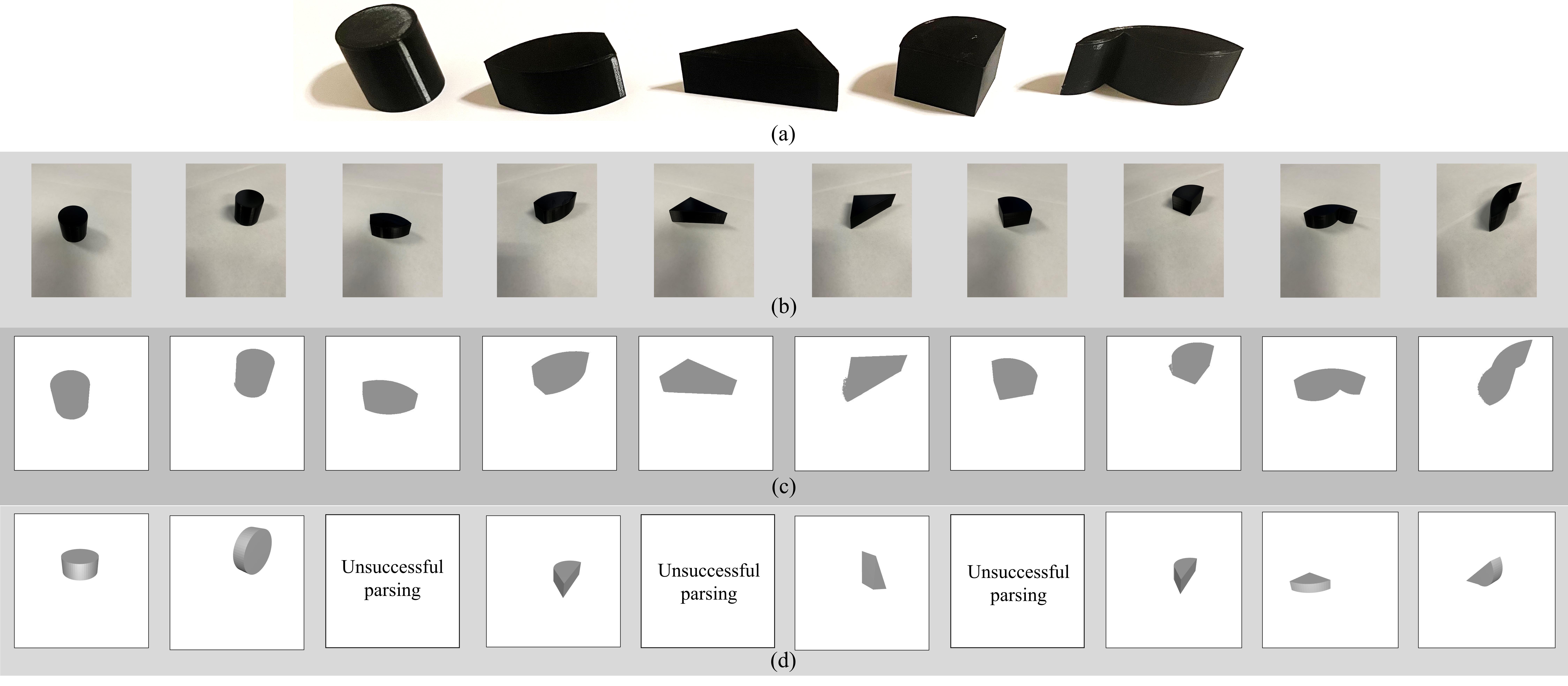}
    \caption{Validation experiments using real-world images. (a) 3D-printed design objects; (b) Photographs (3024$\times$4032 pixels) of the objects taken using a smartphone; (c) Processed images (512$\times$512 pixels) for the input to the Image2CADSeq model; (d) Rendered images of the resultant 3D model from the predicted CAD sequence. Unsuccessful cases are also shown.
}
    \label{fig:validation_3DPrint}
\end{figure*}

\section{DISCUSSION}
\label{sec:discussion}
In this section, we discuss the insights obtained from the experiments as well as the limitations of the current study.

\subsection{THE IMPACT OF DESIGN RULES}
The results in Section \ref{subsec:results} consistently show that the TEA model trained on the dataset with rules outperforms the one without rules. This result underscores the beneficial impact of design rules on the predictive performance of the model in Image2CADSeq tasks.

This has three implications: (1) Compared to the dataset with randomly generated dimension information, the dataset that incorporates design rules introduces latent patterns and relations for the model to learn. 
This suggests that our model is effective in capturing those embedded rules and thus generating more accurate and more realistic designs since real-world CAD models are often created with domain-specific knowledge and design rules. 
(2) The inclusion of design rules can enhance the model's generalizability. It also plays a critical role in minimizing the probability of creating unfeasible CAD designs. (3) Since practical CAD designs often follow domain-specific standards and knowledge, data are therefore inevitably associated with rules. Thus, the proposed method has strong practical implications.

\subsection{TEA VS. TEVAE} 
The improved predictive performance in TEVAE is due to the use of a variational autoencoder (VAE) that can capture the latent design representations of CAD programs. 
The effectiveness of the TEVAE architecture is demonstrated through improved performance across various metrics, significantly exceeding the results achieved in the TEA model. 
The superior performance of the TEVAE model can be largely attributed to the three advantages offered by VAEs.
(1) Unlike traditional AEs, VAEs create a latent space that follows a well-defined and continuous distribution, such as the Gaussian distribution typically used. This design facilitates smoother interpolation between data points, enhancing the capture of meaningful variations in CAD designs. (2) The encoder in a VAE is more efficient in extracting relevant and prominent features from CAD programs than a standard AE. This efficiency stems from the VAE's focus on capturing the underlying data distribution, rather than merely replicating input data. (3) The inclusion of the KL-divergence term in the VAE's loss function helps reduce overfitting. It promotes the model to capture a broader data distribution rather than memorizing specific instances. This enhances TEVAE's generalizability on new, unseen data.

\subsection{LIMITATIONS AND FUTURE WORK}
It is indeed a challenge in our research to further improve the prediction accuracy of operation parameters. 
An important observation from our experiments is the near-perfect reconstruction capability in Stage 1 training with an accuracy of the CAD program (ACP), up to $99.9\%$. However, the problem arises during Stage 2 training, which involves regressing the latent space learned in Stage 1 using images as input. The difficulty lies in aligning the latent representation of the image with that of the CAD programs.
To address this issue, we plan to explore modal alignment techniques as introduced in the recent literature \cite{li2023deep, song2020toward}. They could offer a promising solution to unify different modalities in a single latent space to promote cross-modal synthesis. 

We have been focused on synthesizing simple geometrical shapes, such as cylinders and tri-prisms. While these basic geometries are fundamental to more complex designs, our focus on them has limited the network's capability to handle intricate, real-world design tasks. 
Recognizing this, we acknowledge the need to train the Image2CADSeq model with more diverse and complex datasets to tackle advanced design challenges. 
We plan to collect more sophisticated geometries that mirror the complexities encountered in actual design environments, often embodied as assemblies comprising multiple interconnected components. To achieve this, we plan to explore two primary strategies: (1) Enhancing our data synthesis pipeline: We intend to integrate a wider range of complex geometries into our current data synthesis pipeline. This expansion will allow the network to learn from a broader spectrum of shapes and structures, better preparing it for real-world applications. (2) Using real-world design datasets: Another avenue involves harnessing datasets that include historical CAD modeling process data. An example is the Autodesk Fusion 360 Gallery dataset \cite{willis2021fusion}, which offers a rich source of real-world design examples. Our objective here is to extract CAD sequences that correspond to more intricate designs, similar to the simplified CAD programs outlined in Table \ref{tab:comparison_DSLs}. This approach will enable the network to learn from actual design processes, further enhancing its applicability to practical scenarios.

At the end of this study, we are curious about the model performance in real-world applications where users can take photos of physical artifacts and instantly transform them into CAD sequences. Therefore, we conducted a validation experiment in which five types of template shapes, as listed in Table \ref{tab:template_shapes}, were first 3D printed as shown in Figure \ref{fig:validation_3DPrint} (a). Then, high-resolution photographs (3024$\times$4032 pixels) of these printed objects were captured using a smartphone, as shown in the second row of Figure \ref{fig:validation_3DPrint}. Subsequently, these photographs were resized to 512$\times$512 pixels to facilitate input into the Image2CADSeq model for the generation of CAD sequences. The third row illustrates this preprocessing stage, while the fourth row presents the rendered images of the resultant 3D models, including instances of unsuccessful parsing. The parsing rate achieved in this experiment was 70\%, mirroring the proficiency level indicated in Figure \ref{fig:3Dandimages}. For the 3D models successfully parsed, four were predicted as correct categories but with inaccurate parameters. The remaining three were incorrectly predicted. This indicates that the model performance, when using real-world 3D objects and their image data, is inferior, compared to using the synthesized dataset (i.e., the 3D data generated in CAD) as reported in Section 5.

The result underscores the need for further enhancements in the Image2CADSeq model to improve its accuracy in inferring the CAD representations of images of real-world objects. In particular, enhancing the model's ability to accurately predict parameters is essential to improve the parsing rate. Moreover, to address the model's current limitation in handling various images with different colors, textures, perspectives, or specific artistic styles, 
data augmentation methods, such as the incorporation of objects in different colors and in various lighting conditions or backgrounds, could be beneficial. Such treatments are expected to enrich the model's training data and, thereby, improve its ability to process a wider array of real-world image data.

\section{CONCLUSION}
\label{sec:CONCLUSION}
In this study, we have developed a novel Image2CADSeq model to predict CAD sequences from images. This network, particularly exemplified by the performance of the TEVAE model, aims to revolutionize design methodologies by enabling the conversion of images into operational CAD sequences. A CAD sequence offers more benefits than pure 3D CAD models, such as greater flexibility in modifying CAD operations and managing the historical process/knowledge of CAD model construction.

For training purposes, our focus is on synthesized data representing simple shape primitives. In addition, we propose an evaluation framework that can comprehensively assess model performance. The results obtained are very promising, yet improvement can still be made. 
Therefore, our future efforts will be directed towards (1) Enhancing geometric complexity. We will expand the model's capabilities to encompass a broader spectrum of geometries. This expansion aims to align the model more closely with those in real-world design applications; (2) Incorporating diverse design data. A key area of development involves the integration of more varied and realistic design datasets. This can greatly facilitate the machine learning process; (3) Advancing training methodologies. We plan to explore innovative network architectures and training methodologies to improve the efficiency and adaptability of the model; (4) Incorporating industry standards. Engaging with industry experts will be crucial to guide the development of the model. Their insights will ensure that the model meets practical needs and adheres to industry standards.

In summary, the proposed approach has significant potential to lead to transformative changes in existing CAD systems, revolutionizing the product development cycle. Additionally, it has the potential to promote the democratization of design, allowing people with limited experience or expertise to actively participate in CAD. For example, this approach can help regular customers engage in product design and concept generation, promoting personalized design and creation and human-centered generative design \cite{Demirel2023, li2023deep}. 

\bibliographystyle{asmems4}

\begin{acknowledgment}
The authors gratefully acknowledge the financial support from the National Science Foundation through award 2207408.

\end{acknowledgment}



\end{document}